\documentclass[10pt,twocolumn,letterpaper]{article}

\usepackage{wacv}

\usepackage{amsmath}
\usepackage{amssymb}
\usepackage{amsfonts}
\usepackage{amsopn}
\usepackage{multirow}
\usepackage{booktabs}
\usepackage{array}
\usepackage[ruled,vlined,linesnumbered]{algorithm2e}
\usepackage{subcaption}
\usepackage{makecell}

\providecommand{\stars}[1]{\ensuremath{^{\text{#1}}}}

\definecolor{wacvblue}{rgb}{0.21,0.49,0.74}
\usepackage[pagebackref,breaklinks,colorlinks,allcolors=wacvblue]{hyperref}
\usepackage[accsupp]{axessibility}

\def\wacvPaperID{328}
\def\confName{WACV}
\def\confYear{2027}

\title{Scalable Unrolled Chains for MRI Reconstruction
}

\author{Emiljo M\"ehillaj$^{1}$ and Ender Konukoglu$^{1,2}$\\
$^{1}$Computer Vision Lab, ETH Zurich, Zurich, Switzerland\\
$^{2}$LOOP Zurich - Medical Research Center, Zurich, Switzerland\\
{\tt\small \{emiljo.mehillaj, ender.konukoglu\}@vision.ee.ethz.ch}
}

\begin{document}
\maketitle
\begin{abstract}
Unrolled variational networks achieve state-of-the-art performance for accelerated MRI reconstruction, but standard end-to-end training becomes prohibitively memory intensive as the number and width of cascades increase. 
We propose a memory-efficient training framework that updates a randomly sampled cascade while retaining supervision from the final reconstruction loss. 
The error signal is propagated from the end of the chain to the sampled cascade using an activation-light backward pass that avoids storing full chain activations, thereby decoupling training memory from chain length. 
A time-conditioned network shared across cascades further keeps the parameter count independent of chain length. 
The backward pass is approximate only through frozen normalization statistics, and a trained exact gradient control shows the approximation leaves reconstruction quality unchanged. 
The framework remains trainable at chain lengths and cascade widths where standard end-to-end backpropagation exhausts GPU memory. 
On fastMRI single-coil knee and multi-coil brain reconstruction, our method consistently outperforms other local training approaches. 
On high-SNR knee slices, our method surpasses E2E-VarNet at $8\times$ and matches it at $4\times$, while on the complete brain validation set it surpasses E2E-VarNet at $8\times$ and achieves comparable performance at $4\times$, with fewer parameters, lower peak memory, and faster per-iteration training.
Code is available at \url{https://github.com/emehillaj/scalable-unrolled-mri}.
\end{abstract}
\section{Introduction} \label{sec:intro}

Magnetic Resonance Imaging (MRI) provides excellent soft tissue contrast without ionizing radiation, but its sequential acquisition leads to long scan times and sensitivity to motion.
Accelerated MRI addresses this by undersampling k-space and using reconstruction algorithms to recover diagnostically reliable images from incomplete measurements.
Supervised deep learning, particularly unrolled variational networks~\citep{varnet_hammernik,e2evarnet}, has substantially improved reconstruction quality by repeatedly applying learned refinement and data consistency in a sequence of cascades.
These cascades can be interpreted as successive iterations of a discretized MR reconstruction algorithm~\citep{varnet_hammernik}.
Unrolled networks' capacity can be increased through wider refinement networks or longer chains, with the latter resembling multi-step generative inference procedures such as diffusion models~\citep{ddpm}.
However, end-to-end training stores activations from every cascade, making peak memory scale with both network width and chain depth.
Consequently, scaling either dimension quickly becomes impractical under realistic GPU memory budgets.

Existing approaches trade activation memory for additional computation, architectural constraints, or approximate gradients.
Checkpointing recomputes discarded activations~\citep{chen_sublinearmemorycost}, while reversible, in-place, and computational imaging methods recover them through reversible architectures, locally reconstructible layers, or model-specific inverse operations~\citep{gomez2017revnet,bulo2018inplaceabn,kellman2020memory}.
Other approaches compress activations or modify error propagation or supervision, including activation compression~\citep{chen2021actnn}, synthetic gradients~\citep{jaderberg2017synthetic}, local objectives~\citep{belilovsky2019greedy}, and surrogate or sparsified backpropagation~\citep{chakrabarti2019backprop,sun2017meprop,nokland2016direct}.
These approaches add computation, impose architecture-specific constraints, or approximate the gradient, while decoupled local supervision methods instead modify the training objective by supervising intermediate layers.

Intermediate reconstruction states in unrolled MRI provide natural points for decoupling training across cascades.
Existing approaches decouple learning at these intermediate cascade outputs by defining targets or reconstruction states along a prescribed sequence. 
GLEAM directly supervises intermediate cascade outputs against the fully sampled reference~\citep{gleam}, while k-space cold diffusion progressively modifies the supervising sampling mask from undersampled to fully sampled, and FLAT interpolates between undersampled and fully sampled states to reconstruct intermediate targets~\citep{kcold_diff,flat}.
These approaches avoid or modify full chain backpropagation, but their intermediate targets or prescribed sequences may not align with optimizing the final reconstruction and may therefore constrain the solutions learned by the complete chain.

We propose an efficient sampled training framework for a shared, cascade-conditioned unrolled network that derives its training signal directly from the final reconstruction loss rather than from intermediate objectives.
Unlike synthetic gradients, local learning, and trajectory supervision, our framework retains the final loss signal while computing parameter gradients for only one sampled cascade at each iteration.
The full chain is evaluated forward, with the cascades preceding the sampled one forming a forward only prefix and the remaining cascades forming a frozen suffix.
The suffix propagates the final loss error signal back to the sampled cascade using an activation-light backward pass, where only the sampled cascade performs a standard parameter backward pass.
Thus, the method combines cascade sampling with memory-efficient propagation of the final loss signal through the suffix.

Unlike checkpointing and activation reconstruction, our framework neither recomputes discarded activations nor requires invertible architectures.
Instead, we construct a lightweight backward pass that requires little or no activation storage in the suffix.
For linear layers, the input gradient can be computed directly from the layer weights without storing activations, while for LeakyReLU, a one-bit sign mask is sufficient to recover the exact input gradient.
The only approximation arises at normalization layers, where we hold the forward statistics fixed to linearize the backward map.
Sec.~\ref{sec:exp_fidelity} shows that the resulting error signal closely matches exact suffix backpropagation and leads to indistinguishable training behavior.

Only the sampled cascade performs a standard backward pass and accumulates parameter gradients, while the prefix requires no backward computation and the suffix propagates only input gradients.
This reduces activation storage and backward computation, yielding lower peak memory and faster training steps than standard end-to-end backpropagation (Sec.~\ref{sec:exp_efficiency}).

With independent cascade parameters, uniform cascade sampling amounts to randomized blockwise optimization~\citep{wright2015coordinate}, where each cascade receives fewer updates as the chain length $T$ increases, leading to increasingly suboptimal optimization.
We instead share a single refinement network across cascade positions and condition it on the cascade index using a sinusoidal embedding, following iteration-conditioned unrolled networks and diffusion timestep conditioning~\citep{yun2026time,ddpm}.
This allows every iteration to update parameters reused across the chain while retaining stage-dependent behavior.
We isolate the effect of this parameterization in Sec.~\ref{sec:exp_fidelity}.

We evaluate the framework on fastMRI single-coil knee and multi-coil brain reconstruction at $4\times$ and $8\times$ acceleration.
We compare our method with end-to-end and other memory-efficient training alternatives in terms of reconstruction accuracy, peak training memory, and training time across different chain lengths and network widths (Sec.~\ref{sec:exp_efficiency}).
\section{Method}
\label{sec:method}

\subsection{Problem setup and unrolled reconstruction}

The MRI reconstruction problem of an image $x$ given undersampled k-space measurements $y$ is commonly formulated as:
\begin{equation}
    \hat{x}=\arg\min_x \frac{1}{2}\|Ax-y\|_2^2+\lambda \mathcal{R}(x),
    \label{eq: main_optim}
\end{equation}
where $A$ denotes the MRI encoding operator, including Fourier transform and undersampling mask, 
and $\mathcal{R}$ is a learned or hand-crafted regularizer.

Unrolled networks approximate an iterative solver to the problem in Eq.~\ref{eq: main_optim} by composing $T$ cascades, each consisting of a forward encoding consistency (data consistency) and a learned refinement operator which is generally a convolutional neural network ($\operatorname{CNN}$). 
We denote the intermediate reconstruction after cascade $t$ as $x_t$, with $x_0$ initialized from a zero-filled reconstruction.
Each cascade produces:
\begin{equation}
    {x}_{t+1}={x}_t-\eta_t A^*\left(A{x}_t-y\right)+\operatorname{CNN}_{\theta_t}\left({x}_t\right)
    \label{eqn:updates}
\end{equation}
where $A^*\left(Ax_t-y\right)$ is the gradient of the first term in Eq.~\ref{eq: main_optim}, enforcing consistency with the forward encoding operator and $\eta_t$ is a weighting term. 
$\operatorname{CNN}_{\theta_t}$ with parameters $\theta_t$ is a learnable refinement network (e.g., a U-Net~\citep{unet} operating in image or k-space) enforcing minimization of $\mathcal{R}(x)$.

In end-to-end training, parameters $\{\theta_t\}_{t=0}^{T-1}$ of all cascades are optimized by backpropagating the loss $L(x_T,x_{\mathrm{gt}})$ of the final output $x_T$ with respect to a fully-sampled reference ground truth $x_{\mathrm{gt}}$ through the full chain. 
The exact gradient for cascade $t$ has the form
\begin{equation}
    \frac{\partial L}{\partial \theta_t}=
    \frac{\partial L}{\partial x_{T}}
    \left(\prod_{\ell=T-1}^{t+1}\frac{\partial x_{\ell+1}}{\partial x_{\ell}}\right)
    \frac{\partial x_{t+1}}{\partial \theta_t},
    \label{eq:full_grad}
\end{equation}
where the Jacobian product is taken in order of decreasing $\ell$
(so that $\partial x_T/\partial x_{T-1}$ is leftmost) and is defined to be the identity when $t=T-1$.
This Jacobian product requires storing (or recomputing) intermediate activations for all cascades, leading to memory usage that grows with $T$ (or substantial increases in training time).

\subsection{Memory-efficient backpropagation}
\label{sec:actlight}
In our setting, we backpropagate an approximate error signal stemming from the final reconstruction loss through the entire unrolled chain, while avoiding the activation memory cost of full end-to-end backpropagation. 
At each iteration, we uniformly sample a cascade index $t\in\{0,\dots,T-1\}$ to update.
Cascades $0,\dots,t-1$ form a prefix evaluated without activation storage, cascade $t$ is trained with standard backpropagation starting from the approximate error signal at its output, and cascades $t+1,\dots,T-1$ form a frozen suffix that propagates the loss gradient, i.e., the error signal at the end of the chain, back to $x_{t+1}$.

There are two possible parameterizations of the chain.
In the conventional E2E-VarNet-style parameterization, each cascade has its own distinct network with parameters $\theta_t$.
In our shared time-embedded parameterization, the same network parameters $\theta$ are reused at every cascade, with the cascade index $t$ provided through a time embedding.
Our sampled training applies to both cases: sampling $t$ updates either the corresponding per-cascade block $\theta_t$, or the shared network $\theta$ evaluated at index $t$.
In both cases, only the sampled cascade evaluation contributes parameter gradients.
For notational simplicity, we write the parameters used at cascade $t$ as $\theta_t$, regardless of whether they correspond to a distinct per-cascade network or to the shared time-embedded networks.

Let us assume a cascade $t$ is sampled for training. We denote with $f_{t+1:T-1}$ the frozen suffix, i.e., cascades from $t+1$ to $T-1$, and $f_t$ the sampled cascade. 
In this notation, the sampled cascade outputs $x_{t+1} = f_t(x_t;\theta_t)$ and $f_t$ represents the entire update equation in Equation~\ref{eqn:updates}. The frozen suffix transforms the output of the $t^{\textrm{th}}$ cascade to the prediction as $x_{T} = f_{t+1:T-1}(x_{t+1})$, where for notational simplicity we do not indicate the parameters of the frozen suffix. 
In standard backpropagation, computation of the gradient of the final loss with respect to the parameters $\theta_t$ requires only two entities: (1) $x_t$ and (2) the error signal at the output of the $t^{\textrm{th}}$ cascade. This error signal is computed in terms of the gradient of the loss with respect to the final prediction and the Jacobian of the suffix as:
$$\delta_{t+1} = \frac{\partial L}{\partial x_{t+1}} = \frac{\partial L}{\partial x_T}\frac{\partial x_T}{\partial x_{t+1}} = \frac{\partial L}{\partial x_T}J_{f_{t+1:T-1}}$$
where $J_{f_{t+1:T-1}}$ denotes the Jacobian of $f_{t+1:T-1}$ and $\partial L/\partial x_T$ is the gradient of the final loss with respect to the reconstruction at the end of the chain.
In this setup, computation of $J_{f_{t+1:T-1}}$ is the component that requires memory intensive activation storing for the CNNs. 
Our activation-light backward pass strategy evaluates this component with minimal interventions: it is exact for the linear operators and the LeakyReLU nonlinearities, and approximate only at the normalization layers, whose statistics we hold fixed.
To this end, instead of storing full activations, we store only a lightweight backward state needed to propagate input gradients, motivated by deconvolutional networks ~\citep{zeiler2013}.

CNNs in our cascades are U-Nets~\citep{unet} composed of convolutional and transposed convolutional layers, average pooling, instance normalization, and LeakyReLU nonlinearities.
As in E2E-VarNet, the input to each U-Net is normalized before applying the network.
The suffix $f_{t+1:T-1}$ is effectively a composition of a number of such elementary components and the matrix $J_{f_{t+1:T-1}}$ is the product of Jacobians of successive components due to chain rule.  
We specify the activation-light backward pass for each elementary component and their combination that defines the activation-light backward pass for the entire suffix:

\noindent\textbf{Linear layers:}
Layers such as convolutions, transpose convolutions, and the data consistency layer (forward encoding operator) that apply a linear operator $\mathcal{W}$ to its input,  $z_{\mathrm{out}}=\mathcal{W}z_{\mathrm{in}}$, contribute to the Jacobian computation with their adjoint $\mathcal{W}^*$ (e.g. a transposed convolution for a convolution, and vice versa) as in:
\begin{equation}
 \frac{\partial x_T}{\partial z_{\textrm{in}}} =  \frac{\partial x_T}{\partial z_{\textrm{out}}}\mathcal{W}^*.
\end{equation}
Since $\mathcal{W}^{*}$ depends only on the model weights, which are already in memory, these layers require no cached activation. 
This operation is a linear transformation with known parameters.

\noindent\textbf{LeakyReLU:} 
In standard backpropagation, the input activation to a LeakyReLU~\citep{reluactivation} should be retained so that the backward pass can determine which slope was active at each spatial location.
However, because LeakyReLU is piecewise linear, the full activation values are unnecessary as its backward map depends only on the input sign pattern.
We therefore cache only the binary mask
$\mathcal{B}=(z_{\mathrm{in}}>0)$ of an input $z_{\mathrm{in}}$ to the LeakyReLU and recover the exact input gradient as
\begin{equation}
 \frac{\partial x_T}{\partial z_{\textrm{in}}} = \frac{\partial x_T}{\partial z_{\textrm{out}}}\odot \big(\mathcal{B} + \alpha\,(\mathbf{1}-\mathcal{B})\big),
\end{equation}
where $\alpha$ is the negative slope, $\odot$ denotes elementwise multiplication, $\mathbf{1}$ is the all-ones tensor, and $z_{\textrm{out}}$ in this case refers to the output of the LeakyReLU.
Thus, the full floating-point activation tensor is replaced by a one-bit switch per element.
This is analogous to switch-based backward routing in feature-visualization methods~\citep{zeiler2013,guidedbackprop}, but here the switch recovers the exact LeakyReLU input gradient.

\noindent\textbf{Instance normalization:}
For each sample and channel, instance normalization computes spatial statistics $\mu=\sum_n z_{\textrm{in}}^{(n)}/N$, 
$\sigma^2=\sum_n(z_{\textrm{in}}^{(n)}-\mu)^2/N$, and
$s=\sqrt{\sigma^2+\epsilon}$, with $\epsilon$ a small numerical stability constant, over $N=H\times W$ locations, then outputs $z_{\textrm{out}}^{(n)}=\gamma (z_{\textrm{in}}^{(n)}-\mu)/s+\beta$ with learned affine parameters $\gamma,\beta$.
Due to variance scaling, exact backpropagation depends on the saved normalized activations $\hat z_{\textrm{in}}^{(n)}=(z_{\textrm{in}}^{(n)}-\mu)/s$.
In the activation-light suffix, we freeze the forward statistics and treat the layer as locally affine, yielding the following approximation 

\begin{equation}
  \left(\frac{\partial x_T}{\partial z_{\textrm{in}}}\right)_n\approx {\left(\widehat{\frac{\partial x_T}{\partial z_{\textrm{in}}}}\right)_n} =\left(\frac{\partial x_T}{\partial z_{\textrm{out}}}\right)_n \frac{\gamma}{s}.
  \label{eq:in_approx}
\end{equation}

This stores only $\mathcal{O}(BC)$ statistics for $B$ samples and $C$ channels instead of a full activation tensor.
The same frozen-statistics approximation is applied to the U-Net input standardization. 
Thus, normalization is the only layer type whose input gradient map is approximated in the activation-light suffix backward.

Composing these layer-wise input-gradient maps over each component in a suffix cascade gives an activation-light approximation to the cascade input-output Jacobian
$J_\ell=\partial x_{\ell+1}/\partial x_\ell$ in Eq.~\eqref{eq:full_grad}, which we denote by $\widehat{J}_\ell$.
The resulting gradient estimator for the trained cascade $t$ is
\begin{equation}
    \frac{\partial L}{\partial \theta_t}
    \approx
    \frac{\partial L}{\partial x_T}
    \left(\prod_{\ell=T-1}^{t+1}\widehat{J}_{\ell}\right)
    \frac{\partial x_{t+1}}{\partial \theta_t},
    \label{eq:approx_e2e_grad}
\end{equation}
where the final factor is computed exactly by standard backpropagation through the sampled cascade.
The resulting approximate error signal at the sampled cascade and gradients with respect to parameters of this cascade are then given by 
$$\delta_{t+1} \approx \widehat{\delta}_{t+1} = \frac{\partial L}{\partial x_T}\left(\prod_{\ell=T-1}^{t+1}\widehat{J}_\ell\right)\ \textrm{and}\ \frac{\partial L}{\partial \theta_t}\approx \widehat{\delta}_{t+1}\frac{\partial x_{t+1}}{\partial \theta_t}.$$

The approximate nature of the backward pass stems only from frozen normalization statistics.
Empirically, Sec.~\ref{sec:exp_fidelity} shows that this approximation preserves the gradient direction delivered to the sampled cascade, with only moderate changes in magnitude.
More importantly, it avoids storing or recomputing full activations in the suffix. Therefore, it has computation time comparable to a forward pass through the suffix and is cheaper than standard backpropagation because suffix parameter gradients are not computed.

\subsection{Shared time-conditioned network for depth-scalable training}
\label{sec:timeembed}

In the current standard unrolled reconstruction networks, the refinement CNN networks in different cascades are parameterized as independent networks, whose parameter count grows linearly with the number of cascades $T$.
With independent per-cascade networks, sampled cascade training becomes a block coordinate optimization over $T$ separate parameter blocks.
As $T$ increases, due to sampled cascade training, each block is updated less frequently, making optimization increasingly difficult and potentially leading to suboptimal solutions.

To improve scalability, we instead use a single U-Net shared across cascades and condition it on the cascade index $t$ using a sinusoidal time embedding, analogous to timestep conditioning in diffusion models~\citep{ddpm, kcold_diff, yun2026time}.
The time embedding is injected additively into the U-Net features, allowing one shared parameter set to realize stage-dependent behavior across the reconstruction chain.
This architectural choice has two advantages. 
First, it keeps the number of trainable parameters constant with respect to $T$. 
Second, it converts sampled cascade training from alternating updates over many independent networks into repeated updates of one shared parameter set.
Consequently, every training iteration updates parameters that are reused across the full cascade trajectory, while the time embedding preserves cascade-specific adaptation.
We study the effect of this design empirically in Sec.~\ref{sec:exp_te}.
\section{Experiments}

\label{sec:setup}
\noindent\textbf{Datasets and metrics.}
We evaluated our method against the baselines on the fastMRI single-coil knee and multi-coil brain datasets~\cite{fastmri_dataset,fastmri_dataset_2}, using randomized Cartesian undersampling at acceleration factors of $4\times$ and $8\times$ with center fractions of $8\%$ and $4\%$, respectively.
All methods were trained on the corresponding fastMRI training splits and evaluated on the validation splits.
We report NMSE, PSNR, and SSIM on magnitude images center-cropped according to the fastMRI evaluation protocol~\cite{fastmri_dataset,fastmri_dataset_2}. 
For the multi-coil brain dataset, coil images were combined using root-sum-of-squares (RSS) before metric computation.

Near-empty and very noisy slices skew aggregate metrics, so for the single-coil knee data we split the validation set into high- and low-SNR cohorts using a proxy computed from the fully sampled reference k-space: the ratio of central k-space power to the mean power in the four k-space corners.
We refer to these groups as the high- and low-SNR-proxy cohorts throughout the tables.

All metrics are computed over the complete fastMRI validation splits: $2514$ high-SNR and $4621$ low-SNR knee slices, and all $21{,}842$ brain slices.
Brain results are reported without SNR stratification (Tab.~\ref{tab:main_brain}) because in the multi-coil setting  coil geometry and coil number also affects acquisition quality, so thresholding based on individual coil noise level would select an ill defined subset rather than a noise regime (quantified in the supplementary material).

Extended analyses are deferred to the supplementary material due to space: full volumewise results with bootstrap confidence intervals, unstratified knee metrics, a threshold sensitivity analysis, and cross-acceleration generalization.

\noindent\textbf{Baselines and training.}
We compare against E2E-VarNet~\cite{e2evarnet}, GLEAM~\cite{gleam}, FLAT-style trajectory supervision~\cite{flat}, and K-space Cold Diffusion~\cite{kcold_diff}.
We additionally include E2E-VarNet with activation checkpointing~\cite{chen_sublinearmemorycost} as a memory reduction baseline, which buys reduced activation memory at the expense of additional compute.
Its memory and time are quantified in Sec.~\ref{sec:exp_cost}.

All unrolled methods use an E2E-VarNet-style backbone~\cite{e2evarnet} with $T$ cascades and U-Net architecture.
The U-Net has four pooling stages and starts with $C$ base channels, doubling the channel width after each downsampling stage.
Complex MR images are split into real and imaginary channels and all U-Net convolutions are standard real-valued convolutions over this two-channel representation.
Our method and K-space Cold Diffusion add a diffusion-style time embedding to the U-Net ~\cite{ddpm, kcold_diff}. 
FLAT-style trajectory supervision uses the velocity-alignment loss between an interpolated initialization and the fully sampled reference. 
We deliberately isolate this local loss rather than reproducing the full FLAT-style objective, whose additional E2E term requires full chain backpropagation and therefore carries E2E-VarNet's activation memory cost, placing it outside the memory-efficient regime compared here. 
We evaluate the trajectory supervision alone, under the same budget as the other baselines, and write ``FLAT-style'' throughout.

All unrolled methods are trained on the fastMRI single-coil knee or multi-coil brain training set with MSE loss using AdamW~\cite{adamw}, learning rate $2{\times}10^{-5}$ and an effective batch size of $8$ (gradient accumulation).
K-space Cold Diffusion uses its original $\ell_1$ diffusion objective~\cite{kcold_diff}.

\noindent\textbf{Multi-coil extension.} 
For multi-coil data, our method and the unrolled baselines follow the E2E-VarNet formulation~\cite{e2evarnet}. 
A sensitivity network estimates coil sensitivity maps from the acquired k-space data. 
The coil images are then SENSE-combined before U-Net processing, and the resulting image is expanded back to the individual coils using the same maps before multi-coil data consistency is applied~\cite{sense_pruessman}. 
Since K-space Cold Diffusion was originally designed for single-coil data, we equip it with the same sensitivity estimation and coil combination operators and perform its diffusion process in the coil-combined image domain.

\subsection{Training scheme analysis}
\label{sec:exp_fidelity}

\noindent\textbf{Error signal fidelity.}
First, we verified that the activation-light error signal obtained from our custom backward pass approximates the true suffix error signal
obtained from a standard autograd backward pass. 
On single-coil knee at $T{=}8$, we compare activation-light and exact autograd error signals on $200$ high-SNR slices across the seven cascade positions.

The median cosine similarity was $0.996$, and $99\%$ of the paired signals had cosine similarity above $0.91$.
The relative $\ell_2$ error was $5$--$12\%$ (interquartile range).

To calibrate what a perturbation of this size means for optimization, we compare it against the stochastic gradient noise the optimizer already absorbs. 
Replacing the exact suffix backward by the activation-light one rotated the parameter update by a median of $12.3^\circ$, whereas drawing a different minibatch rotated the exact update by $86.0^\circ$.
The approximation is therefore roughly $7\times$ smaller than the sampling noise stochastic optimization already tolerates.
Full distributions, the depth sweep, the per suffix-length breakdown and the corresponding results for the independently parameterized model are given in the supplementary material.

\noindent\textbf{Sharing, sampling, and approximation.}
We next isolate the contributions of parameter sharing, cascade sampling, and activation-light suffix propagation. Tab.~\ref{tab:isolate} compares matched configurations at $T{=}8$ and $C{=}32$.

Parameter sharing with full backpropagation achieved $29.68$\,dB, compared with $29.54$\,dB for E2E-VarNet with distinct cascade parameters. 
Replacing full backpropagation with sampled training reduced PSNR by $0.11$\,dB ($29.68\rightarrow29.57$). 
Although the cascade weights are shared, each iteration uses the gradient contribution from one sampled cascade rather than aggregating contributions from all $T$ positions. 
Under uniform sampling, this is an unbiased but higher variance estimate of the full gradient. 
A larger batch size in the sampled scheme could reduce this sampling variance,
although we do not investigate this effect here. 
Sampling nevertheless reduced per iteration training time from $224$ to $136$\,ms.

Replacing the exact suffix gradient with the activation-light approximation left reconstruction quality unchanged ($29.57\rightarrow29.58$\,dB), while reducing peak training memory from $6.7$ to $1.1$\,GB and per iteration time from $136$ to $134$\,ms. Thus, sampling provides the main computational saving, while activation-light propagation provides the main memory saving.

\label{sec:exp_te}
\noindent\textbf{Effect of time embedding.}
Isolating the shared time-embedded U-Net within the sampled framework, the shared network outperformed independent per-cascade U-Nets at matched depth and width ($29.58$ vs.\ $29.24$\ dB at $T{=}8$ and $29.70$ vs.\ $29.10$ at $T{=}15$, high-SNR knee $8\times$) with $8$-$14\times$ fewer parameters and a count independent of depth.
Because the shared network is updated at every step rather than only when its block is sampled, it also avoids the block-coordinate instability of the per-cascade regime; the full comparison is tabulated in the supplementary material.

\begin{table}[t]
\centering
\setlength{\tabcolsep}{1.5pt}
\footnotesize
\caption{\textbf{Isolating the training scheme.} Single-coil knee $8\times$, $T{=}8$, $C{=}32$; each row changes one design decision. PSNR on the complete high-SNR cohort ($n{=}2514$). $\pm$ is seed noise from three separately trained replicates.}
\label{tab:isolate}
\begin{tabular}{l c cc}
\toprule
Training scheme & PSNR$\uparrow$ & \makecell{Mem\\(GB)$\downarrow$} & \makecell{Time\\(ms)}\\
\midrule
E2E-VarNet (distinct, full BP) & 29.54 & 5.9 & 215\\
\midrule
Shared-TE, full BP & 29.68\,\tiny$\pm$0.02 & 6.7 & 224\\
Shared-TE, sampled, exact suffix & 29.57\,\tiny$\pm$0.02 & 6.7 & 136\\
Shared-TE, sampled, activation-light (ours) & 29.58\,\tiny$\pm$0.02 & 1.1 & 134\\
\bottomrule
\end{tabular}%

\end{table}

\subsection{Main comparison}
\label{sec:exp_main}
\begin{table*}[t]
\centering\footnotesize
\setlength{\tabcolsep}{3pt}
\caption{\textbf{Main comparison on single-coil knee data at $T{=}15$.} Results are reported as NMSE/PSNR/SSIM, with per slice standard deviations below, for high-SNR ($n{=}2514$) and low-SNR ($n{=}4621$) validation cohorts defined using the k-space SNR proxy. 
Columns {$>$}35 / {$\leq$}35 are the high-/low-SNR-proxy cohorts. Best per cohort in \textbf{bold}; $C$ = base channel width, last column = trainable parameters.
For Ours (96), the second line is the fraction of slices beating E2E-VarNet (32); stars are results of one-sided paired Wilcoxon tests comparing the two, Holm-Bonferroni corrected over the 18 tests reported in Tabs.~\ref{tab:main} and~\ref{tab:main_brain}. 
At these cohort sizes a starred low-SNR cell can differ from the reference by less than the displayed precision, so the win fraction and the effect size, not the star, carry the practical claim. Slice-level tests treat slices as independent and therefore ignore within-volume correlation; the volume-level analysis in the supplementary confirms the PSNR and NMSE conclusions at the volume unit (${*}\,p<0.05$, ${**}\,p<0.01$, ${***}\,p<0.001$);}

\begin{tabular}{lcccc c}
\toprule
& \multicolumn{2}{c}{$4\times$} & \multicolumn{2}{c}{$8\times$} & \multirow{2}{*}{Params (M)} \\
\cmidrule(lr){2-3}\cmidrule(lr){4-5}
Method ($C$) & \makecell{High-SNR\\proxy {$>$}35} & \makecell{Low-SNR\\proxy {$\leq$}35} & \makecell{High-SNR\\proxy {$>$}35} & \makecell{Low-SNR\\proxy {$\leq$}35} & \\
\midrule
E2E-VarNet (32)
& \makecell{0.0094/32.41/0.790\\[-1pt]\scriptsize(0.0068/2.71/0.081)}
& \makecell{0.0574/26.83/0.538\\[-1pt]\scriptsize(0.0448/3.08/0.132)}
& \makecell{0.0160/29.90/0.716\\[-1pt]\scriptsize(0.0091/2.34/0.093)}
& \makecell{0.0691/25.79/0.450\\[-1pt]\scriptsize(0.0470/2.88/0.138)}
& 116.4 \\
\addlinespace[2pt]
E2E + ckpt. (64)
& \makecell{\textbf{0.0093}/32.42/\textbf{0.791}\\[-1pt]\scriptsize(0.0066/2.72/0.081)}
& \makecell{0.0578/26.82/0.538\\[-1pt]\scriptsize(0.0459/3.09/0.132)}
& \makecell{0.0154/30.07/0.716\\[-1pt]\scriptsize(0.0090/2.39/0.095)}
& \makecell{0.0699/25.74/0.446\\[-1pt]\scriptsize(0.0474/2.88/0.138)}
& 465.4 \\
\addlinespace[2pt]
E2E + ckpt. (96)
& \makecell{\textbf{0.0093}/\textbf{32.46}/\textbf{0.791}\\[-1pt]\scriptsize(0.0066/2.73/0.081)}
& \makecell{0.0572/26.84/0.539\\[-1pt]\scriptsize(0.0447/3.08/0.132)}
& \makecell{0.0153/30.13/0.717\\[-1pt]\scriptsize(0.0090/2.40/0.096)}
& \makecell{0.0699/25.75/0.444\\[-1pt]\scriptsize(0.0480/2.89/0.138)}
& 1047.1 \\
\addlinespace[2pt]
GLEAM (32)
& \makecell{0.0103/31.94/0.777\\[-1pt]\scriptsize(0.0068/2.60/0.082)}
& \makecell{0.0597/26.64/0.521\\[-1pt]\scriptsize(0.0466/3.05/0.137)}
& \makecell{0.0181/29.33/0.700\\[-1pt]\scriptsize(0.0094/2.27/0.094)}
& \makecell{0.0730/25.51/0.435\\[-1pt]\scriptsize(0.0484/2.84/0.138)}
& 116.4 \\
\addlinespace[2pt]
FLAT-style (32)
& \makecell{0.0208/28.62/0.688\\[-1pt]\scriptsize(0.0092/2.16/0.073)}
& \makecell{0.0928/24.88/0.487\\[-1pt]\scriptsize(0.0838/3.02/0.109)}
& \makecell{0.0424/25.50/0.576\\[-1pt]\scriptsize(0.0165/2.18/0.085)}
& \makecell{0.1320/23.11/0.361\\[-1pt]\scriptsize(0.1073/2.82/0.108)}
& 116.4 \\
\addlinespace[2pt]
K-space Cold Diff.~(64)
& \makecell{0.0100/31.91/0.788\\[-1pt]\scriptsize(0.0053/2.46/0.070)}
& \makecell{\textbf{0.0505}/\textbf{27.21}/\textbf{0.611}\\[-1pt]\scriptsize(0.0342/2.96/0.083)}
& \makecell{0.0174/29.46/0.707\\[-1pt]\scriptsize(0.0085/2.27/0.089)}
& \makecell{\textbf{0.0661}/\textbf{25.89}/\textbf{0.491}\\[-1pt]\scriptsize(0.0401/2.80/0.108)}
& 32.3 \\
\midrule
Ours (32)
& \makecell{0.0095/32.32/0.787\\[-1pt]\scriptsize(0.0067/2.69/0.081)}
& \makecell{0.0580/26.78/0.534\\[-1pt]\scriptsize(0.0453/3.07/0.134)}
& \makecell{0.0167/29.70/0.703\\[-1pt]\scriptsize(0.0093/2.33/0.095)}
& \makecell{0.0737/25.53/0.435\\[-1pt]\scriptsize(0.0512/2.87/0.136)}
& \textbf{8.1} \\
\addlinespace[2pt]
Ours (64)
& \makecell{0.0096/32.28/0.787\\[-1pt]\scriptsize(0.0067/2.68/0.081)}
& \makecell{0.0587/26.75/0.531\\[-1pt]\scriptsize(0.0465/3.08/0.138)}
& \makecell{0.0153/30.10/0.713\\[-1pt]\scriptsize(0.0088/2.40/0.096)}
& \makecell{0.0724/25.66/0.437\\[-1pt]\scriptsize(0.0524/2.92/0.140)}
& 32.3 \\
\addlinespace[2pt]
Ours (96)
& \makecell{\textbf{0.0093}\stars{***}/32.44\stars{***}/\textbf{0.791}\stars{***}\\[-1pt]\scriptsize(0.0066/2.72/0.081)\\[-1pt]\scriptsize 78/78/81\% wins}
& \makecell{0.0575/26.83\stars{***}/0.537\\[-1pt]\scriptsize(0.0451/3.08/0.135)\\[-1pt]\scriptsize 53/53/34\% wins}
& \makecell{\textbf{0.0146}\stars{***}/\textbf{30.33}\stars{***}/\textbf{0.721}\stars{***}\\[-1pt]\scriptsize(0.0085/2.43/0.095)\\[-1pt]\scriptsize 95/95/74\% wins}
& \makecell{0.0695/25.81\stars{***}/0.446\\[-1pt]\scriptsize(0.0489/2.92/0.140)\\[-1pt]\scriptsize 50/50/15\% wins}
& 72.7 \\
\bottomrule
\end{tabular}
\label{tab:main}
\end{table*}

\begin{table*}[t]
\centering\small
\setlength{\tabcolsep}{6pt}
\caption{\textbf{Main comparison on multi-coil brain data at $T{=}15$.} per slice NMSE/PSNR/SSIM are reported over the full validation set ($n{=}21{,}842$), with standard deviations below. Best results per metric are in \textbf{bold}.
Win fractions, stars, and conventions as in Tab.~\ref{tab:main} (same 18-test Holm family).
$^\dagger$FLAT-style did not optimize on brain: its outputs stay within $0.5\%$ of the zero-filled input and its loss decreased by ${<}0.02\%$ over two days of training. We report the row rather than omitting it; the same implementation trains normally on knee.}
\label{tab:main_brain}
\begin{tabular}{l cc c}
\toprule
Method ($C$) & $4\times$ & $8\times$ & Params (M)\\
\midrule
E2E-VarNet (32)
& \makecell{0.0058/37.42/0.915\\[-1pt]\scriptsize(0.0067/2.96/0.096)}
& \makecell{0.0106/34.38/0.890\\[-1pt]\scriptsize(0.0075/2.64/0.102)}
& 116.8 \\
\addlinespace[2pt]
GLEAM (32)
& \makecell{0.0081/35.76/0.895\\[-1pt]\scriptsize(0.0081/2.72/0.098)}
& \makecell{0.0180/32.11/0.863\\[-1pt]\scriptsize(0.0178/2.71/0.102)}
& 116.8 \\
\addlinespace[2pt]
FLAT-style (32)$^\dagger$
& \makecell{0.0514/27.62/0.750\\[-1pt]\scriptsize(0.0358/2.81/0.080)}
& \makecell{0.1100/24.07/0.631\\[-1pt]\scriptsize(0.0565/2.58/0.084)}
& 116.8 \\
\addlinespace[2pt]
K-space Cold Diff.\ (64)
& \makecell{0.0173/32.57/0.851\\[-1pt]\scriptsize(0.0242/3.05/0.102)}
& \makecell{0.0284/29.90/0.799\\[-1pt]\scriptsize(0.0285/2.85/0.109)}
& 32.3 \\
\midrule
Ours (32)
& \makecell{0.0063/37.03/0.912\\[-1pt]\scriptsize(0.0072/2.93/0.099)}
& \makecell{0.0112/34.16/0.887\\[-1pt]\scriptsize(0.0081/2.64/0.103)}
& \textbf{8.6} \\
\addlinespace[2pt]
Ours (64)
& \makecell{\textbf{0.0056}/37.64/0.917\\[-1pt]\scriptsize(0.0067/3.01/0.097)}
& \makecell{0.0105/34.49/0.891\\[-1pt]\scriptsize(0.0087/2.72/0.104)}
& 32.8 \\
\addlinespace[2pt]
Ours (96)
& \makecell{\textbf{0.0056}\stars{***}/\textbf{37.71}\stars{***}/\textbf{0.917}\stars{***}\\[-1pt]\scriptsize(0.0069/3.05/0.096)\\[-1pt]\scriptsize 93/93/95\% wins}
& \makecell{\textbf{0.0096}\stars{***}/\textbf{34.91}\stars{***}/\textbf{0.894}\stars{***}\\[-1pt]\scriptsize(0.0077/2.72/0.104)\\[-1pt]\scriptsize 93/93/93\% wins}
& 73.2 \\
\bottomrule
\end{tabular}
\end{table*}

We first compare all methods at fixed depth $T{=}15$ on the single-coil knee task, across both $4\times$ and $8\times$ acceleration and high and low SNR cohorts (Tab.~\ref{tab:main}).
Among the memory-efficient training methods, our shared time-embedded model was consistently strongest on the high-SNR knee cohorts.
At matched width $C{=}32$, the proposed method reached $29.70$\,dB on high-SNR knee with $8\times$ acceleration, outperforming GLEAM ($29.33$\,dB), FLAT-style trajectory supervision ($25.50$\,dB), and K-space Cold Diffusion ($29.46$\,dB), while having only $8.1$M trainable parameters.
Increasing the shared network width further improved reconstruction quality.

Compared with E2E-VarNet (with 32 channels), which uses independent per-cascade networks and $116.4$M parameters at $T{=}15$, our method exhibited two regimes.
At matched width $C{=}32$, our approach trailed E2E-VarNet by a small margin on high-SNR knee ($29.70$ vs.\ $29.90$\,dB at $8\times$, and $32.32$ vs.\ $32.41$\,dB at $4\times$), while using roughly $14\times$ fewer parameters
and substantially lower peak training memory ($1.1$\,GB vs.\ $11.0$\,GB).
When part of this saved budget was reinvested into a wider shared network, the proposed method surpassed E2E-VarNet on knee $8\times$ ($30.10$\,dB at $C{=}64$ and $30.33$\,dB at $C{=}96$, compared with $29.90$\,dB for E2E-VarNet).
At knee $4\times$, our approach with $C{=}96$ matched E2E-VarNet ($32.44$ vs.\ $32.41$\,dB). Even in these cases, both the number of model parameters and the memory requirements were lower than E2E-VarNet. 

Tab.~\ref{tab:main} also reports E2E-VarNet trained with activation checkpointing at $C{=}64$ and $C{=}96$. 
Since checkpointed gradients are exact, these rows train the same models as E2E-VarNet at those widths and tracked our variants closely: $30.07$ and $30.13$\,dB at $8\times$ high-SNR against our $30.10$ and $30.33$, the $C{=}96$ arm reaching $32.46$\,dB at $4\times$, slightly above our $32.44$. 
This quality, however, cost $2$-$2.5\times$ our step time and an order of magnitude more parameters (Sec.~\ref{sec:exp_cost}).

On low-SNR knee cohorts, the differences between methods were smaller.
K-space Cold Diffusion was competitive, particularly on knee $4\times$ and $8\times$ low-SNR slices, while the proposed method remained close to or above E2E-VarNet as width increased.
These observations support the noise-sensitivity analysis in the supplementary material: scaling capacity is most beneficial on cleaner, high-SNR slices where anatomical detail dominates.

\begin{figure}[h]
    \centering
    \begin{subfigure}[t]{\linewidth}
        \centering
        \includegraphics[width=0.94\linewidth]{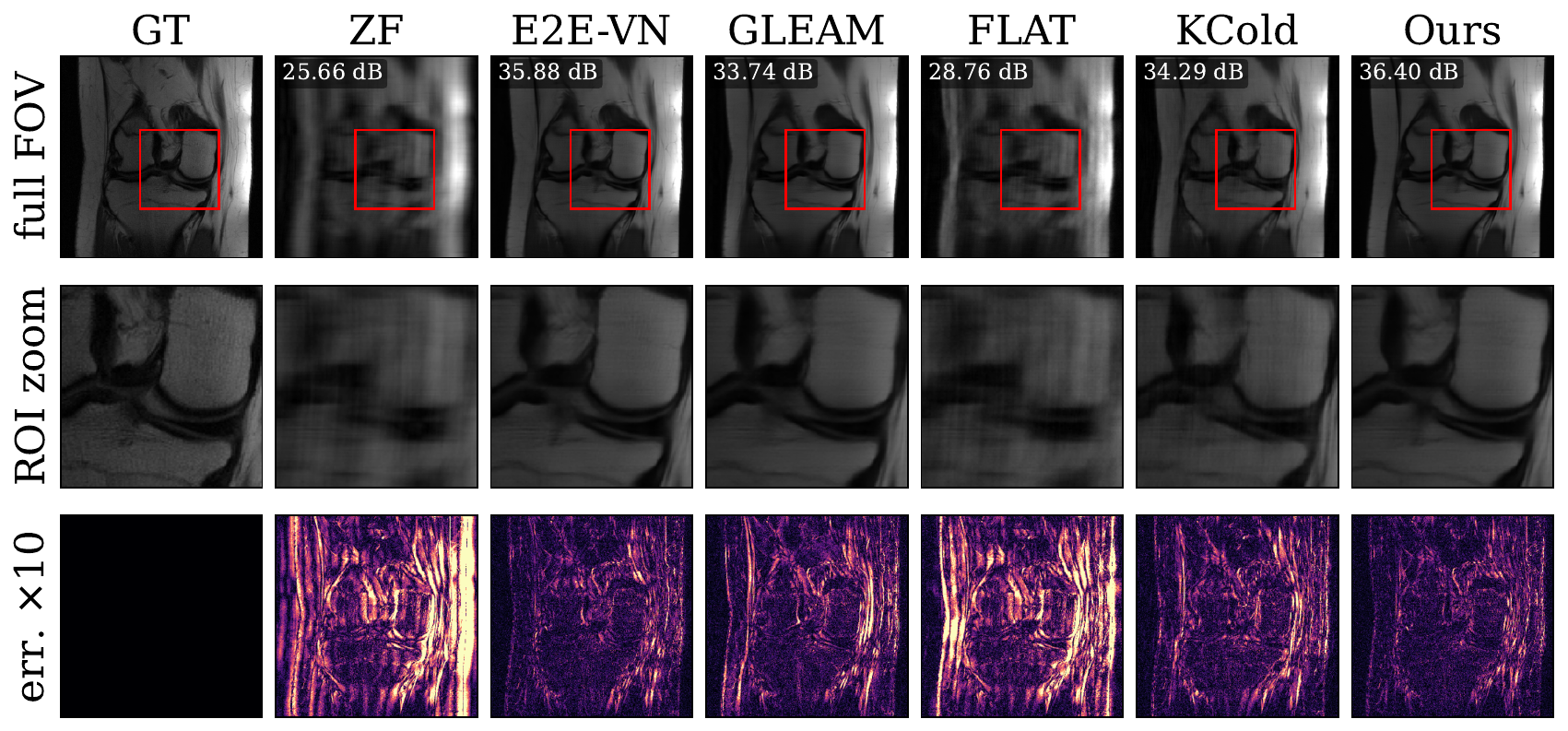}
        \caption{Single-coil knee.}
        \label{fig:recon_knee}
    \end{subfigure}\\[3pt]
    \begin{subfigure}[t]{\linewidth}
        \centering
        \includegraphics[width=0.94\linewidth]{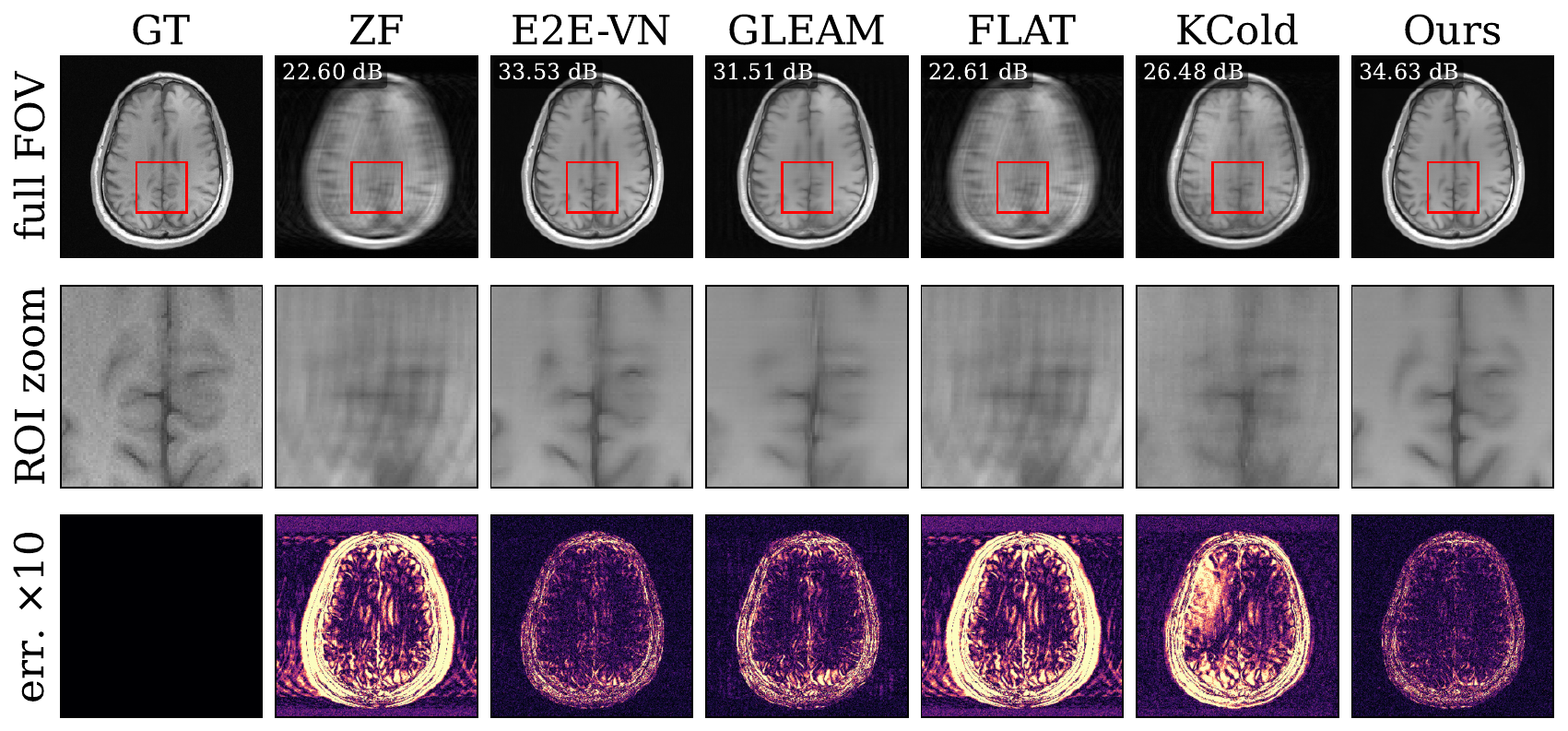}
        \caption{Multi-coil brain, RSS magnitude.}
        \label{fig:recon_brain}
    \end{subfigure}
    \caption{\textbf{Qualitative comparison at $8\times$ acceleration.}
    One validation slice per anatomy, each rendered as three rows: full FOV (ROI in red, PSNR inset), ROI zoom, and error $|\hat{x}-x_{\mathrm{ref}}|$ ($\times10$).
    Columns: fully sampled reference, zero-filled (ZF) input, E2E-VarNet (E2E-VN) (32), GLEAM, FLAT-style, K-space Cold Diffusion (KCold), and Ours (96).
    }
    \label{fig:reconstructions}
\end{figure}

Fig.~\ref{fig:reconstructions} shows two knee $8\times$ slices where the proposed method recovered fine anatomical structure with residual error comparable to E2E-VarNet and visibly lower than GLEAM, FLAT-style, and K-space Cold Diffusion.

On multi-coil brain, reported without SNR stratification (see Setup), Tab.~\ref{tab:main_brain} indicates that the same capacity argument held, and the crossover arrived earlier than on knee.
At matched width $C{=}32$ our method stayed within $0.39$\,dB of E2E-VarNet at $4\times$ ($37.03$ vs.\ $37.42$\,dB) and $0.22$\,dB at $8\times$ ($34.16$ vs.\ $34.38$\,dB) while training $13.6\times$ fewer parameters ($8.6$\,M vs.\ $116.8$\,M).
Reinvesting the saved memory into width closed and then reversed that gap already at $C{=}64$ ($37.64$ and $34.49$\,dB), and at $C{=}96$ our method attained the best value of every metric at both accelerations ($37.71$ and $34.91$\,dB).
The margin is larger at $8\times$ than at $4\times$, consistent with capacity mattering most where the inverse problem is least constrained.
Additional qualitative examples for both anatomies are provided in the supplementary material.

\label{sec:exp_noise}

\subsection{Computational efficiency}
\label{sec:exp_cost}
\label{sec:exp_efficiency}

\noindent Sampled activation-light training reshapes the computational trade-off along four axes: peak training memory, parameter count, per iteration training time, and inference cost.
Tab.~\ref{tab:cascade} reports the first three, together with reconstruction quality on the high-SNR knee cohort, for E2E-VarNet, its checkpointed variant, and our method.
Depth is swept at fixed width $C{=}32$ (top block) and width at fixed depth $T{=}8$ (bottom block).
Quality improved with both depth and width over these ranges, so we concentrate here on the accompanying computational costs.
All measurements use a single NVIDIA RTX~A6000 ($48$\,GB), batch size $1$, and $640{\times}372$ resolution, in full fp32; the peak memory convention is stated in Sec.~4.2 of the supplementary material and applied identically to every method.

\begin{table}[t]
\centering
\setlength{\tabcolsep}{1.5pt}
\footnotesize
\caption{\textbf{Capacity scaling}, single-coil knee $8\times$, high-SNR cohort (SNR${>}35$\,dB): depth at fixed $C{=}32$ (top) and width at fixed $T{=}8$ (bottom). Quality on the complete high-SNR cohort ($n{=}2514$); Mem/Time on one A6000 at batch $1$.}
\label{tab:cascade}
\label{tab:width}
\begin{tabular}{l c ccc cc c}
\toprule
Method & $T$/$C$ & NMSE$\downarrow$ & PSNR$\uparrow$ & SSIM$\uparrow$ & \makecell{Mem\\(GB)$\downarrow$} & \makecell{Time\\(ms)} & \makecell{Params\\(M)}\\
\midrule
\multicolumn{8}{l}{\emph{Depth scaling} ($C{=}32$, sweep $T$)}\\
E2E-VarNet & 8   & 0.0173 & 29.54 & 0.708 & 5.9 & 215 & 62.05\\
E2E-VarNet & 15  & 0.0160 & 29.90 & 0.716 & 11.0 & 403 & 116.35\\
E2E + ckpt. & 8  & 0.0173 & 29.54 & 0.708 & 1.3 & 293 & 62.05\\
E2E + ckpt. & 15 & 0.0160 & 29.90 & 0.716 & 1.7 & 551 & 116.35\\
E2E + ckpt. & 25 & 0.0156 & 30.02 & 0.715 & 2.3 & 901 & 193.9 \\
Ours & 8   & 0.0172 & 29.58 & 0.701 & 1.1 & 134 & 8.08\\
Ours & 15  & 0.0167 & 29.70 & 0.703 & 1.1 & 231 & 8.08\\
Ours & 25  & 0.0162 & 29.86 & 0.707 & 1.2 & 364 & 8.08\\
Ours & 50  & 0.0156 & 30.04 & 0.711 & 1.3 & 719 & 8.08\\
\midrule
\multicolumn{8}{l}{\emph{Width scaling} ($T{=}8$, sweep $C$)}\\
E2E-VarNet & 32  & 0.0173 & 29.54 & 0.708 & 5.9 & 215 & 62.05\\
E2E + ckpt. & 64 & 0.0158 & 29.95 & 0.714 & 3.4 & 602 & 248.2 \\
E2E + ckpt. & 96 & 0.0154 & 30.08 & 0.717 & 6.4 & 1089 & 558.4 \\
Ours & 32  & 0.0172 & 29.58 & 0.701 & 1.1 & 134 & 8.08\\
Ours & 64  & 0.0159 & 29.93 & 0.711 & 2.3 & 277 & 32.3\\
Ours & 96  & 0.0153 & 30.10 & 0.718 & 3.5 & 485 & 72.7\\
Ours & 128 & 0.0151 & 30.15 & 0.718 & 4.8 & 685 & 129.3\\
\bottomrule
\end{tabular}
\end{table}

\noindent\textbf{Training memory.}
Measured peak training memory of our method grew only marginally with chain length ($1.1$\,GB at $T{=}8$ to $1.3$\,GB at $T{=}50$), whereas E2E-VarNet's grew approximately linearly in $T$ ($5.9$ to $36.5$\,GB) and exceeded the $48$\,GB budget beyond $T{\approx}64$. Width scaling behaved the same way: E2E-VarNet's memory grew steeply and exceeded $48$\,GB, while we trained at $C{=}160$ under $7$\,GB. The corresponding scaling curves are plotted in the supplementary material.
A shared-TE backbone trained with full backpropagation tracks E2E-VarNet's memory and also runs out, so the saving comes from activation handling, not the architecture.

\noindent\textbf{Parameters.}
Reusing one network at every cascade makes parameter count independent of depth: $8.1$\,M at any $T$, against $62$\,M ($T{=}8$) to $776$\,M ($T{=}100$) for per-cascade methods. Even our $C{=}96$ variant ($72.7$\,M) stays below the $T{=}15$ baselines.

\noindent\textbf{Training time.}
Per iteration training time grew along both scaling axes but stayed below full backpropagation in either direction, because most of the chain never computes parameter gradients: they are computed only for the sampled cascade (Sec.~3.1 of the supplementary material decomposes the saving by backward component).

Along chain length (Tab.~\ref{tab:cascade}), per iteration time grew because both the forward pass and the activation-light suffix traversal lengthen with $T$ (from $134$\ ms at $T{=}8$ to $719$\ ms at $T{=}50$), while still remaining cheaper than E2E-VarNet at matched depth ($231$ vs.\ $403$\ ms at $T{=}15$).
Along width (Tab.~\ref{tab:cascade}, width block), per iteration time grew as well because each cascade's convolutions scale roughly quadratically in $C$, so per step cost rises even though peak memory rises only modestly.

\noindent\textbf{Inference cost.}
With no backward pass at inference, our method and E2E-VarNet both reduce to $T$ forward cascade evaluations independent of the training scheme: at $C{=}32$ one cascade costs ${\approx}8.8$\,ms, giving ${\approx}131$\,ms at $T{=}15$. 
K-space Cold Diffusion's $T{=}125$-step sampler cost ${\approx}2.2$\,s per slice, ${\approx}17\times$ slower at lower quality (Tab.~\ref{tab:main}).

\noindent\textbf{Comparison with activation checkpointing.}
Checkpointing recomputes activations for exact gradients, trading time for memory.
At $T{=}8$, it reduces memory from $5.9$ to $1.3$\,GB but increases step time from $215$ to $293$\,ms, whereas our method requires $1.1$\,GB and $134$\,ms.
Memory nevertheless increases from $1.3$ to $6.8$\,GB over $T{=}8{\rightarrow}100$, compared with $1.1$-$1.5$\,GB for ours, and parameter count grows with depth because each cascade has a distinct network ($776$\,M at $T{=}100$ vs.\ our constant $8.1$\,M).
Because checkpointing retains E2E-VarNet's distinct per-cascade networks and exact gradients, any higher quality at matched depth or width reflects its substantially larger parameter count.
This comes at roughly $2$-$2.5\times$ our step time and an order of magnitude more parameters.
Larger scale behavior is discussed in the supplementary material.

\section{Conclusion}
\label{sec:conclusion}

We introduced a memory-efficient training framework for scaling unrolled MRI reconstruction in chain length and network width, while retaining supervision from the final reconstruction loss. 
Each iteration updates a single sampled cascade by standard backpropagation and propagates an activation-light approximation of the final loss error through the remaining cascades. 
With shared, time-conditioned cascade weights, parameter count remains independent of chain length, while peak training memory stays nearly constant as the chain grows.

Experiments on fastMRI knee and brain show that our method enables substantially longer and wider chains under fixed memory budgets without sacrificing quality. 
At matched depth, our widest model surpasses E2E-VarNet at $8\times$ and matches it at $4\times$ with fewer parameters, while outperforming other memory-efficient baselines. 
Although computation increases with chain length and width, it remains lower than full end-to-end training at matched depth. 
We also find that scaling gains depend on data quality, motivating SNR-stratified evaluation when studying capacity scaling.

Our contribution is a training strategy rather than a new architecture. 
We demonstrate it with a plain E2E-VarNet-style backbone, leaving stronger backbones as a natural direction for improving reconstruction quality. 
Our experiments use fastMRI knee and brain data, and since the method makes no dataset-specific assumptions, evaluation on other raw k-space datasets is a natural next step.
Overall, the framework enables scaling unrolled reconstruction beyond the memory limits of conventional training and may extend to volumetric and other high dimensional settings.

\section*{Acknowledgments}
This work is partially funded by the Swiss National Science Foundation (SNSF) under grant No. 205320\_200877.
{
    \small
    \bibliographystyle{ieeenat_fullname}
    \bibliography{main}
}

\clearpage
\section*{Supplementary Material}
\label{sec:supp}

\section{Extended quantitative results}
\subsection{Slice-level win fractions (single-coil knee)}
\label{sec:supp_slice_paired}
Tab.~\ref{tab:main} reports the win fraction against the reference row only for Ours~($C{=}96$).
Tab.~\ref{tab:slice_paired_knee} gives it for every method, on the same slices and under the same protocol.
The win fraction measures how consistently a method leads rather than by how much, which matters most where the means are close.
At $8\times$ on the high-SNR cohort Ours~($C{=}96$) beats E2E-VarNet on $95\%$ of the slices and Ours~($C{=}64$) on $70\%$.
On the low-SNR cohort Ours~($C{=}96$) wins on $50\%$, so the two are interchangeable there, in line with the noise sensitivity analysis of the main paper.
K-space Cold Diffusion is the only baseline that beats the reference on the low-SNR cohorts ($67\%$ at $8\times$, $85\%$ at $4\times$), where its denoising style sampler benefits from a reference that is itself noisy.

\begin{table}[b]
\centering\footnotesize
\setlength{\tabcolsep}{3pt}
\caption{\textbf{Slice-level win fractions, single-coil knee.}
Same slices and protocol as the knee block of Tab.~\ref{tab:main}, whose cohort means the table reproduces.
Each cell gives the mean PSNR (dB) and, beneath it, the fraction of slices on which the method beats the reference row (E2E-VarNet, $C{=}32$).
Best value per acceleration is indicated in \textbf{bold}.
}
\label{tab:slice_paired_knee}
\begin{tabular}{l cc cc}
\toprule
& \multicolumn{2}{c}{$4\times$} & \multicolumn{2}{c}{$8\times$}\\
\cmidrule(lr){2-3}\cmidrule(lr){4-5}
Method & high-SNR & low-SNR & high-SNR & low-SNR\\
\midrule
E2E-VarNet (32) & 32.41 & 26.83 & 29.90 & 25.79\\ \addlinespace[2pt]
GLEAM (32) & \makecell{31.94\\{\scriptsize 0\% wins}} & \makecell{26.64\\{\scriptsize 0\% wins}} & \makecell{29.33\\{\scriptsize 4\% wins}} & \makecell{25.51\\{\scriptsize 1\% wins}}\\ \addlinespace[2pt]
FLAT-style (32) & \makecell{28.62\\{\scriptsize 0\% wins}} & \makecell{24.88\\{\scriptsize 0\% wins}} & \makecell{25.50\\{\scriptsize 0\% wins}} & \makecell{23.11\\{\scriptsize 0\% wins}}\\ \addlinespace[2pt]
K-space Cold Diff.\ (64) & \makecell{31.91\\{\scriptsize 22\% wins}} & \makecell{\textbf{27.21}\\{\scriptsize 85\% wins}} & \makecell{29.46\\{\scriptsize 25\% wins}} & \makecell{\textbf{25.89}\\{\scriptsize 67\% wins}}\\ \addlinespace[2pt]
Ours (32) & \makecell{32.32\\{\scriptsize 4\% wins}} & \makecell{26.78\\{\scriptsize 4\% wins}} & \makecell{29.70\\{\scriptsize 21\% wins}} & \makecell{25.53\\{\scriptsize 2\% wins}}\\ \addlinespace[2pt]
Ours (64) & \makecell{32.28\\{\scriptsize 1\% wins}} & \makecell{26.75\\{\scriptsize 0\% wins}} & \makecell{30.10\\{\scriptsize 70\% wins}} & \makecell{25.66\\{\scriptsize 15\% wins}}\\ \addlinespace[2pt]
Ours (96) & \makecell{\textbf{32.44}\\{\scriptsize 78\% wins}} & \makecell{26.83\\{\scriptsize 53\% wins}} & \makecell{\textbf{30.33}\\{\scriptsize 95\% wins}} & \makecell{25.81\\{\scriptsize 50\% wins}}\\
\bottomrule
\end{tabular}%

\end{table}

\subsection{Volumewise, unstratified evaluation with confidence intervals}
\label{sec:supp_volumewise}
Tabs.~\ref{tab:vol_brain8x}, \ref{tab:vol_brain4x}, \ref{tab:vol_knee8x} and~\ref{tab:vol_knee4x} report the volumewise evaluation on the full fastMRI validation split on both anatomies and both accelerations: brain uses all $1{,}378$ validation volumes and knee all $199$.
Beneath each entry we report the fraction of volumes on which the method beats the reference row (E2E-VarNet, $C{=}32$).

On knee our capacity scaled model attains the best PSNR and NMSE at $8\times$ and ties the reference in SSIM, while at $4\times$ our model and the E2E-VarNet rows lie within $0.03$\,dB of one another.
On brain our approach attains the best volumewise PSNR, SSIM and NMSE at both accelerations.

To quantify the margin over the reference, we compute for each volume the PSNR difference between Ours~($C{=}96$) and E2E-VarNet ($C{=}32$), and average these paired differences across volumes.
We obtain a $95\%$ confidence interval by resampling the volumes with replacement $10{,}000$ times and taking the $2.5$th and $97.5$th percentiles of the resulting mean differences.
We report each result as the mean difference followed by its $95\%$ confidence interval: $+0.62$\,dB $[+0.61,+0.63]$ at brain $8\times$ (win $100\%$), $+0.32$\,dB $[+0.31,+0.33]$ at brain $4\times$ (win $99\%$), $+0.21$\,dB $[+0.17,+0.25]$ at knee $8\times$ (win $83\%$), and $+0.016$\,dB $[+0.013,+0.019]$ at knee $4\times$ (win $78\%$).
These intervals quantify uncertainty in the average margin across volumes, rather than the variability of the individual volumetric differences.

\begin{table}[t]
\centering\footnotesize
\caption{\textbf{Volumewise evaluation, multi-coil brain $8\times$.}
All $V{=}1{,}378$ validation volumes, no subsampling.
Each cell reports the metric averaged over volumes and, beneath it, the fraction of volumes on which the method beats the reference row (E2E-VarNet (32)).
Best value per metric is indicated in \textbf{bold}.
}
\label{tab:vol_brain8x}
\begin{tabular}{l ccc}
\toprule
Method & PSNR$\uparrow$ & SSIM$\uparrow$ & NMSE$\downarrow$\\
\midrule
E2E-VarNet (32) & 36.75 & 0.9350 & 0.0093\\ \addlinespace[2pt]
GLEAM (32) & \makecell{34.31\\{\scriptsize 0\% wins}} & \makecell{0.9122\\{\scriptsize 0\% wins}} & \makecell{0.0160\\{\scriptsize 0\% wins}}\\ \addlinespace[2pt]
FLAT-style (32) & \makecell{25.91\\{\scriptsize 0\% wins}} & \makecell{0.6852\\{\scriptsize 0\% wins}} & \makecell{0.1074\\{\scriptsize 0\% wins}}\\ \addlinespace[2pt]
K-space Cold Diff.\ (64) & \makecell{30.78\\{\scriptsize 4\% wins}} & \makecell{0.8313\\{\scriptsize 2\% wins}} & \makecell{0.0351\\{\scriptsize 2\% wins}}\\ \addlinespace[2pt]
Ours (32) & \makecell{36.44\\{\scriptsize 2\% wins}} & \makecell{0.9320\\{\scriptsize 0\% wins}} & \makecell{0.0099\\{\scriptsize 2\% wins}}\\ \addlinespace[2pt]
Ours (64) & \makecell{36.92\\{\scriptsize 81\% wins}} & \makecell{0.9365\\{\scriptsize 84\% wins}} & \makecell{0.0089\\{\scriptsize 81\% wins}}\\ \addlinespace[2pt]
Ours (96) & \makecell{\textbf{37.37}\\{\scriptsize 100\% wins}} & \makecell{\textbf{0.9390}\\{\scriptsize 100\% wins}} & \makecell{\textbf{0.0081}\\{\scriptsize 100\% wins}}\\
\bottomrule
\end{tabular}
\end{table}

\begin{table}[t]
\centering\footnotesize
\caption{\textbf{Volumewise evaluation, multi-coil brain $4\times$.}
All $V{=}1{,}378$ volumes, with the conventions of Tab.~\ref{tab:vol_brain8x}.}
\label{tab:vol_brain4x}
\begin{tabular}{l ccc}
\toprule
Method & PSNR$\uparrow$ & SSIM$\uparrow$ & NMSE$\downarrow$\\
\midrule
E2E-VarNet (32) & 40.02 & 0.9546 & 0.0044\\ \addlinespace[2pt]
GLEAM (32) & \makecell{38.24\\{\scriptsize 0\% wins}} & \makecell{0.9397\\{\scriptsize 0\% wins}} & \makecell{0.0065\\{\scriptsize 0\% wins}}\\ \addlinespace[2pt]
FLAT-style (32) & \makecell{29.47\\{\scriptsize 0\% wins}} & \makecell{0.7974\\{\scriptsize 0\% wins}} & \makecell{0.0477\\{\scriptsize 0\% wins}}\\ \addlinespace[2pt]
K-space Cold Diff.\ (64) & \makecell{32.92\\{\scriptsize 2\% wins}} & \makecell{0.8731\\{\scriptsize 2\% wins}} & \makecell{0.0221\\{\scriptsize 2\% wins}}\\ \addlinespace[2pt]
Ours (32) & \makecell{39.50\\{\scriptsize 0\% wins}} & \makecell{0.9520\\{\scriptsize 0\% wins}} & \makecell{0.0049\\{\scriptsize 0\% wins}}\\ \addlinespace[2pt]
Ours (64) & \makecell{40.26\\{\scriptsize 99\% wins}} & \makecell{0.9557\\{\scriptsize 99\% wins}} & \makecell{0.0042\\{\scriptsize 99\% wins}}\\ \addlinespace[2pt]
Ours (96) & \makecell{\textbf{40.34}\\{\scriptsize 99\% wins}} & \makecell{\textbf{0.9560}\\{\scriptsize 99\% wins}} & \makecell{\textbf{0.0041}\\{\scriptsize 99\% wins}}\\
\bottomrule
\end{tabular}
\end{table}

\begin{table}[t]
\centering\footnotesize
\caption{\textbf{Volumewise evaluation, single-coil knee $8\times$.}
All $V{=}199$ volumes, with the conventions of Tab.~\ref{tab:vol_brain8x}.
}
\label{tab:vol_knee8x}
\begin{tabular}{l ccc}
\toprule
Method & PSNR$\uparrow$ & SSIM$\uparrow$ & NMSE$\downarrow$\\
\midrule
E2E-VarNet (32) & 31.09 & \textbf{0.6770} & 0.0409\\ \addlinespace[2pt]
E2E + ckpt.\ (64) & 31.17 & 0.6758 & 0.0408\\ \addlinespace[2pt]
E2E + ckpt.\ (96) & 31.18 & 0.6759 & 0.0408\\ \addlinespace[2pt]
GLEAM (32) & \makecell{30.68\\{\scriptsize 0\% wins}} & \makecell{0.6651\\{\scriptsize 0\% wins}} & \makecell{0.0440\\{\scriptsize 0\% wins}}\\ \addlinespace[2pt]
FLAT-style (32) & \makecell{27.75\\{\scriptsize 0\% wins}} & \makecell{0.5578\\{\scriptsize 0\% wins}} & \makecell{0.0771\\{\scriptsize 0\% wins}}\\ \addlinespace[2pt]
K-space Cold Diff.\ (64) & \makecell{29.45\\{\scriptsize 21\% wins}} & \makecell{0.6653\\{\scriptsize 39\% wins}} & \makecell{0.0498\\{\scriptsize 8\% wins}}\\ \addlinespace[2pt]
Ours (32) & \makecell{30.87\\{\scriptsize 5\% wins}} & \makecell{0.6656\\{\scriptsize 0\% wins}} & \makecell{0.0430\\{\scriptsize 5\% wins}}\\ \addlinespace[2pt]
Ours (64) & \makecell{31.12\\{\scriptsize 36\% wins}} & \makecell{0.6697\\{\scriptsize 8\% wins}} & \makecell{0.0415\\{\scriptsize 36\% wins}}\\ \addlinespace[2pt]
Ours (96) & \makecell{\textbf{31.30}\\{\scriptsize 83\% wins}} & \makecell{\textbf{0.6770}\\{\scriptsize 45\% wins}} & \makecell{\textbf{0.0401}\\{\scriptsize 83\% wins}}\\
\bottomrule
\end{tabular}
\end{table}

\begin{table}[t]
\centering\footnotesize
\caption{\textbf{Volumewise evaluation, single-coil knee $4\times$.}
All $V{=}199$ volumes, with the conventions of Tab.~\ref{tab:vol_brain8x}.}
\label{tab:vol_knee4x}
\begin{tabular}{l ccc}
\toprule
Method & PSNR$\uparrow$ & SSIM$\uparrow$ & NMSE$\downarrow$\\
\midrule
E2E-VarNet (32) & 32.75 & 0.7387 & 0.0310\\ \addlinespace[2pt]
E2E + ckpt.\ (64) & \textbf{32.78} & 0.7396 & \textbf{0.0309}\\ \addlinespace[2pt]
E2E + ckpt.\ (96) & \textbf{32.78} & \textbf{0.7397} & \textbf{0.0309}\\ \addlinespace[2pt]
GLEAM (32) & \makecell{32.45\\{\scriptsize 0\% wins}} & \makecell{0.7283\\{\scriptsize 0\% wins}} & \makecell{0.0324\\{\scriptsize 0\% wins}}\\ \addlinespace[2pt]
FLAT-style (32) & \makecell{30.09\\{\scriptsize 0\% wins}} & \makecell{0.6632\\{\scriptsize 0\% wins}} & \makecell{0.0490\\{\scriptsize 0\% wins}}\\ \addlinespace[2pt]
K-space Cold Diff.\ (64) & \makecell{30.51\\{\scriptsize 14\% wins}} & \makecell{0.7340\\{\scriptsize 40\% wins}} & \makecell{0.0367\\{\scriptsize 15\% wins}}\\ \addlinespace[2pt]
Ours (32) & \makecell{32.69\\{\scriptsize 1\% wins}} & \makecell{0.7360\\{\scriptsize 1\% wins}} & \makecell{0.0313\\{\scriptsize 1\% wins}}\\ \addlinespace[2pt]
Ours (64) & \makecell{32.65\\{\scriptsize 1\% wins}} & \makecell{0.7346\\{\scriptsize 1\% wins}} & \makecell{0.0316\\{\scriptsize 1\% wins}}\\ \addlinespace[2pt]
Ours (96) & \makecell{32.77\\{\scriptsize 78\% wins}} & \makecell{0.7385\\{\scriptsize 50\% wins}} & \makecell{\textbf{0.0309}\\{\scriptsize 78\% wins}}\\
\bottomrule
\end{tabular}
\end{table}

\subsection{Ablations and capacity scaling under volumewise aggregation}
\label{sec:supp_abl_vol}
The ablations and the capacity scaling of the main paper are evaluated on the complete high-SNR cohort ($n{=}2514$ slices).
To check that none of their conclusions depends on per slice aggregation or on the cohort selection, Tab.~\ref{tab:abl_vol} repeats every ablation under the unstratified volumewise protocol of Sec.~\ref{sec:supp_volumewise}, on all $199$ validation volumes.
Every ordering is preserved.
Quality increases monotonically with depth and with width.

Independent per-cascade networks remain behind the shared time-embedded network and, unlike it, degrade when the chain is lengthened ($-0.10$\,dB from $T{=}8$ to $T{=}15$, against $+0.10$\,dB for the shared network), as predicted by the vanishing per-block update frequency of block-coordinate training.

The training scheme ladder keeps its order as well.
Absolute differences are compressed relative to the per slice protocol (the depth ladder spans $0.30$\,dB here against $0.46$\,dB per slice), because volumewise PSNR normalizes by the volume maximum rather than by the slice maximum, so the two protocols agree in ordering rather than in magnitude.

\begin{table}[t]
\centering\footnotesize
\caption{\textbf{Ablations under volumewise, unstratified evaluation.}
Single-coil knee $8\times$, all $V{=}199$ validation volumes, no SNR selection, aggregated as in Sec.~\ref{sec:supp_volumewise}.
The training scheme rows carry the training budget caveat of Tab.~\ref{tab:isolate} and are the mean over 3 seeds.}
\label{tab:abl_vol}
\begin{tabular}{l ccc}
\toprule
Configuration & PSNR$\uparrow$ & SSIM$\uparrow$ & NMSE$\downarrow$\\
\midrule
\multicolumn{4}{l}{\emph{Depth} ($C{=}32$)}\\
\quad Ours, $T{=}8$ & 30.77 & 0.6629 & 0.0439\\
\quad Ours, $T{=}15$ & 30.87 & 0.6656 & 0.0430\\
\quad Ours, $T{=}25$ & 30.96 & 0.6677 & 0.0427\\
\quad Ours, $T{=}50$ & 31.07 & 0.6698 & 0.0419\\
\midrule
\multicolumn{4}{l}{\emph{Width} ($T{=}8$)}\\
\quad Ours, $C{=}32$ & 30.77 & 0.6629 & 0.0439\\
\quad Ours, $C{=}64$ & 31.02 & 0.6706 & 0.0418\\
\quad Ours, $C{=}96$ & 31.19 & 0.6770 & 0.0404\\
\quad Ours, $C{=}128$ & 31.21 & 0.6765 & 0.0404\\
\midrule
\multicolumn{4}{l}{\emph{Network sharing} ($C{=}32$)}\\
\quad Distinct, $T{=}8$ & 30.59 & 0.6616 & 0.0447\\
\quad Distinct, $T{=}15$ & 30.49 & 0.6591 & 0.0456\\
\quad Shared-TE, $T{=}8$ & 30.77 & 0.6629 & 0.0439\\
\quad Shared-TE, $T{=}15$ & 30.87 & 0.6656 & 0.0430\\
\midrule
\multicolumn{4}{l}{\emph{Training scheme} ($T{=}8$, $C{=}32$)}\\
\quad Shared-TE, full BP & 30.81 & 0.6642 & 0.0436\\
\quad Shared-TE, sampled, exact suffix & 30.77 & 0.6630 & 0.0439\\
\quad Shared-TE, sampled, act.-light & 30.77 & 0.6629 & 0.0439\\
\bottomrule
\end{tabular}%

\end{table}

\subsection{Cross-acceleration generalization}
\label{sec:exp_xaccel}
We also examine how the trained models transfer across acceleration factors.
Each single-coil knee model trained at $8\times$ is evaluated directly at $4\times$, a denser sampling than seen in training, and compared with the same architecture trained at $4\times$ (Tab.~\ref{tab:xaccel}, high-SNR cohort).
Every method loses quality under this shift.
Among the unrolled methods our widest shared model loses the least: at $C{=}96$ it transfers to $32.12$\,dB, $0.32$\,dB below its native $4\times$ result, against drops of $0.45$\,dB for E2E-VarNet and $0.52$\,dB for GLEAM.
K-space Cold Diffusion drops least of all ($-0.17$\,dB), as expected of an iterative sampler whose construction does not depend on the acceleration, but it does so from a lower absolute level.
Our matched-width model ($C{=}32$) drops most ($-0.77$\,dB), so the robustness belongs to the capacity-scaled model rather than to the training scheme alone.

\begin{table}[t]
\centering
\setlength{\tabcolsep}{4pt}
\footnotesize
\caption{\textbf{Cross-acceleration generalization}, single-coil knee, high-SNR cohort (SNR${>}35$\,dB, complete cohort, $n{=}2514$).
All values are PSNR (dB), $\Delta$ is the transfer gap, and the best value per column is in \textbf{bold}.}
\label{tab:xaccel}
\begin{tabular}{l ccc}
\toprule
Method ($C$) & Native $4\times$ & $8\times\!\to\!4\times$ & $\Delta$\\
\midrule
E2E-VarNet (32)          & 32.41 & 31.96 & $-0.45$\\
GLEAM (32)               & 31.94 & 31.42 & $-0.52$\\
FLAT-style (32)          & 28.62 & 28.21 & $-0.41$\\
K-space Cold Diff.\ (64) & 31.91 & 31.74 & $\mathbf{-0.17}$\\
\midrule
Ours (32)                & 32.32 & 31.55 & $-0.77$\\
Ours (64)                & 32.28 & 31.91 & $-0.37$\\
Ours (96)                & \textbf{32.44} & \textbf{32.12} & $-0.32$\\
\bottomrule
\end{tabular}
\end{table}

\section{Noise sensitivity and the SNR proxy}
\subsection{Noise sensitivity and pooled evaluation (single-coil knee)}
\label{sec:supp_pooled}
\begin{figure*}[t]
\centering
\includegraphics[width=0.7\textwidth]{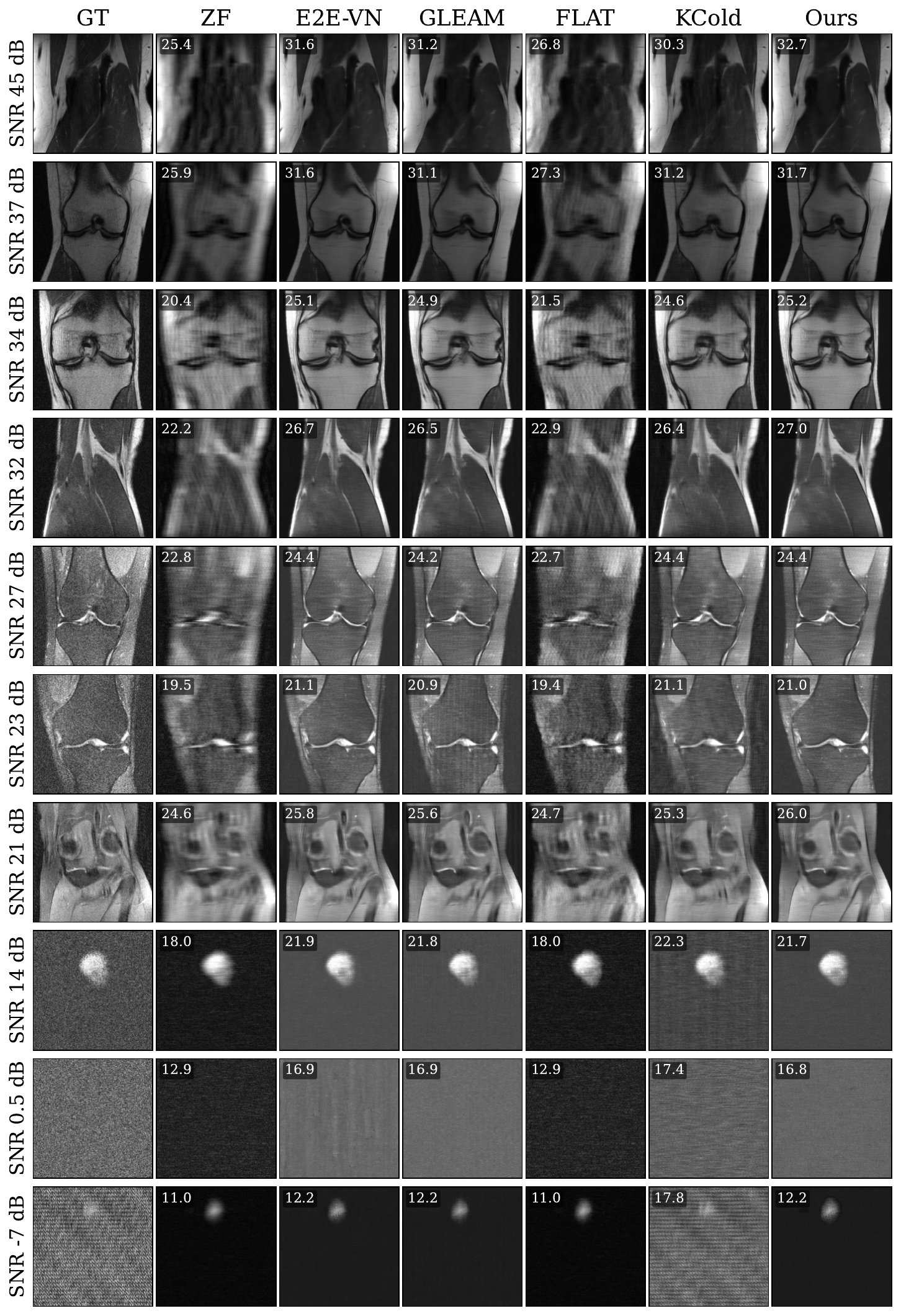}
\caption{\textbf{Reconstruction quality across the SNR range, single-coil knee $8\times$.}
Rows are validation slices ordered by the SNR of the fully sampled reference, from $45$\,dB (top) to $-6.7$\,dB (bottom), with the PSNR (dB) of each method inset.
Columns follow Fig.~\ref{fig:reconstructions}.}
\label{fig:snr_ladder}
\end{figure*}

Tab.~\ref{tab:pooled} reports single-coil knee results on the complete fastMRI validation set ($n{=}7135$ slices), pooled across all SNRs rather than split into the cohorts of Tab.~\ref{tab:main}.
The pooled average is dominated by low-SNR slices, where denoising-style reconstruction is favored.
Accordingly, K-space Cold Diffusion attains the best pooled NMSE and SSIM at both accelerations and the best pooled PSNR at $4\times$, while our widest shared model attains the best pooled PSNR at $8\times$.
This agrees with the SNR-stratified analysis (Tab.~\ref{tab:main} and Sec.~\ref{sec:exp_noise} of the main paper): the ranking of the methods depends strongly on the SNR regime, and capacity scaling of the shared network pays off most on the high-SNR cohort, where anatomical detail rather than noise dominates the residual.
Fig.~\ref{fig:snr_ladder} illustrates this across the full SNR range at $8\times$.
When the fully sampled reference is clean, the metrics are reliable and our capacity scaled model reconstructs the anatomy with the highest fidelity.
As the SNR decreases the reference itself becomes dominated by noise, so PSNR, SSIM and NMSE increasingly reward agreement with that noise rather than with the anatomy.
Our method visibly denoises and preserves structure, yet is penalized for departing from the noisy reference, whereas the iterative sampler of K-space Cold Diffusion stays closer to the noisy reference and is scored higher.
The effect is already visible at intermediate SNR, where the diffusion sampler retains residual structured noise and our reconstruction is cleaner and closer to the underlying anatomy.
Figs.~\ref{fig:noise_knee} and ~\ref{fig:noise_knee4x} plot the same effect as mean PSNR per SNR cohort on both anatomies and both accelerations.
The unrolled methods keep their order in every bin, and K-space Cold Diffusion overtakes them only in the noisiest bins.

\begin{table}[t]
\centering
\setlength{\tabcolsep}{4pt}
\footnotesize
\caption{\textbf{Pooled validation (no SNR stratification), single-coil knee.}
NMSE$\downarrow$/PSNR$\uparrow$/SSIM$\uparrow$ on the complete validation set, pooled across all SNRs ($n{=}7135$), at $4\times$ and $8\times$.
Per slice standard deviation in parentheses, best value per metric and acceleration in \textbf{bold}.}
\label{tab:pooled}
\resizebox{\columnwidth}{!}{%
\begin{tabular}{l cc c}
\toprule
Method ($C$) & Knee $4\times$ & Knee $8\times$ & Params (M)\\
\midrule
E2E-VarNet (32) & \makecell{0.0405/28.79/0.627\\{\scriptsize(0.0429/3.98/0.168)}} & \makecell{0.0504/27.24/0.543\\{\scriptsize(0.0458/3.34/0.178)}} & 116.4\\
\addlinespace[2pt]
GLEAM (32) & \makecell{0.0423/28.51/0.612\\{\scriptsize(0.0445/3.85/0.172)}} & \makecell{0.0537/26.86/0.528\\{\scriptsize(0.0473/3.22/0.177)}} & 116.4\\
\addlinespace[2pt]
FLAT-style (32) & \makecell{0.0674/26.20/0.558\\{\scriptsize(0.0759/3.28/0.137)}} & \makecell{0.1004/23.95/0.437\\{\scriptsize(0.0969/2.85/0.144)}} & 116.4\\
\addlinespace[2pt]
K-space Cold Diff.\ (64) & \makecell{\textbf{0.0362}/\textbf{28.87}/\textbf{0.673}\\{\scriptsize(0.0338/3.59/0.116)}} & \makecell{\textbf{0.0489}/27.14/\textbf{0.567}\\{\scriptsize(0.0401/3.13/0.145)}} & 32.3\\
\midrule
Ours (32) & \makecell{0.0409/28.74/0.623\\{\scriptsize(0.0434/3.96/0.169)}} & \makecell{0.0536/27.00/0.529\\{\scriptsize(0.0497/3.35/0.178)}} & \textbf{8.1}\\
\addlinespace[2pt]
Ours (64) & \makecell{0.0420/28.67/0.622\\{\scriptsize(0.0462/3.97/0.169)}} & \makecell{0.0523/27.23/0.534\\{\scriptsize(0.0505/3.47/0.182)}} & 32.3\\
\addlinespace[2pt]
Ours (96) & \makecell{0.0405/28.81/0.626\\{\scriptsize(0.0432/3.99/0.170)}} & \makecell{0.0502/\textbf{27.41}/0.543\\{\scriptsize(0.0476/3.50/0.182)}} & 72.7\\
\bottomrule
\end{tabular}}
\end{table}

\subsection{Sensitivity to the SNR proxy threshold}
\label{sec:supp_thresh}
\begin{figure}[t]
\centering
\includegraphics[width=\linewidth]{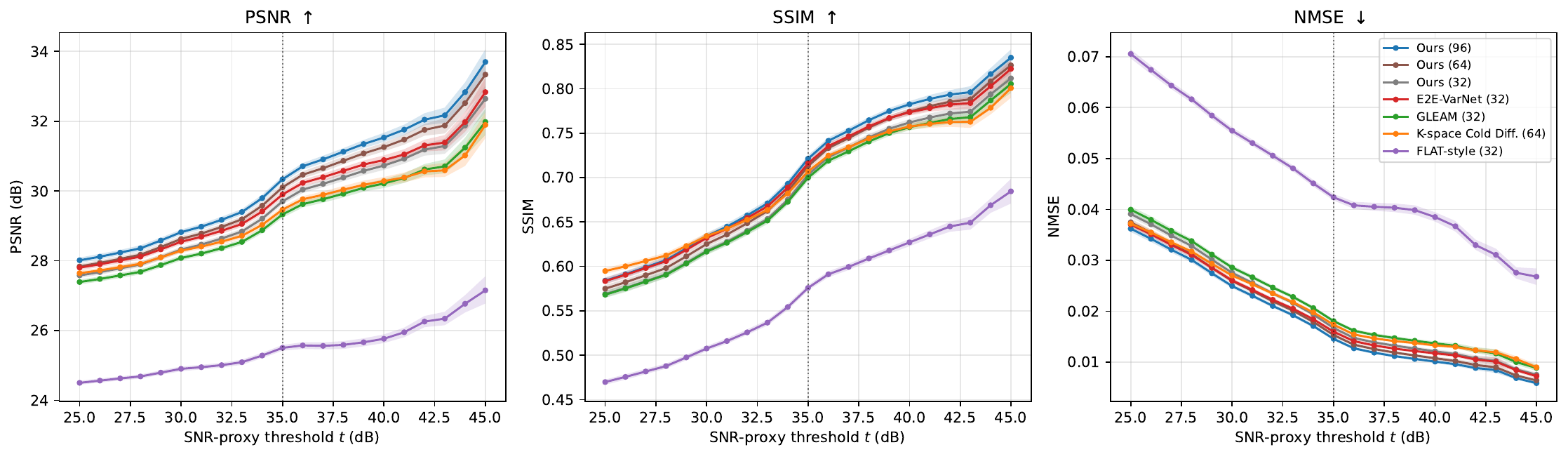}
\caption{\textbf{SNR threshold sensitivity, single-coil knee $8\times$.}
Shaded bands are $95\%$ bootstrap intervals of the cohort mean and the dashed line marks the $35$\,dB cutoff of the main paper.
The rankings are stable over the full sweep.}
\label{fig:thresh_knee8}
\end{figure}

\begin{figure}[t]
\centering
\includegraphics[width=\linewidth]{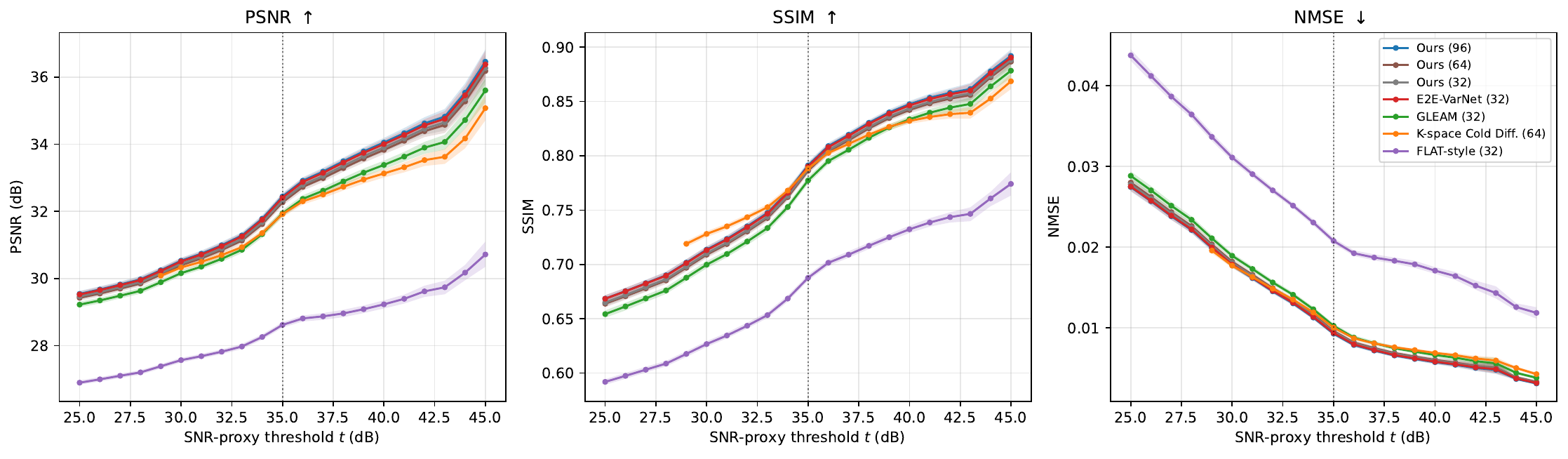}
\caption{\textbf{SNR threshold sensitivity, single-coil knee $4\times$.} Conventions as in Fig.~\ref{fig:thresh_knee8}.}
\label{fig:thresh_knee4}
\end{figure}

\begin{figure}[t]
\centering
\includegraphics[width=\linewidth]{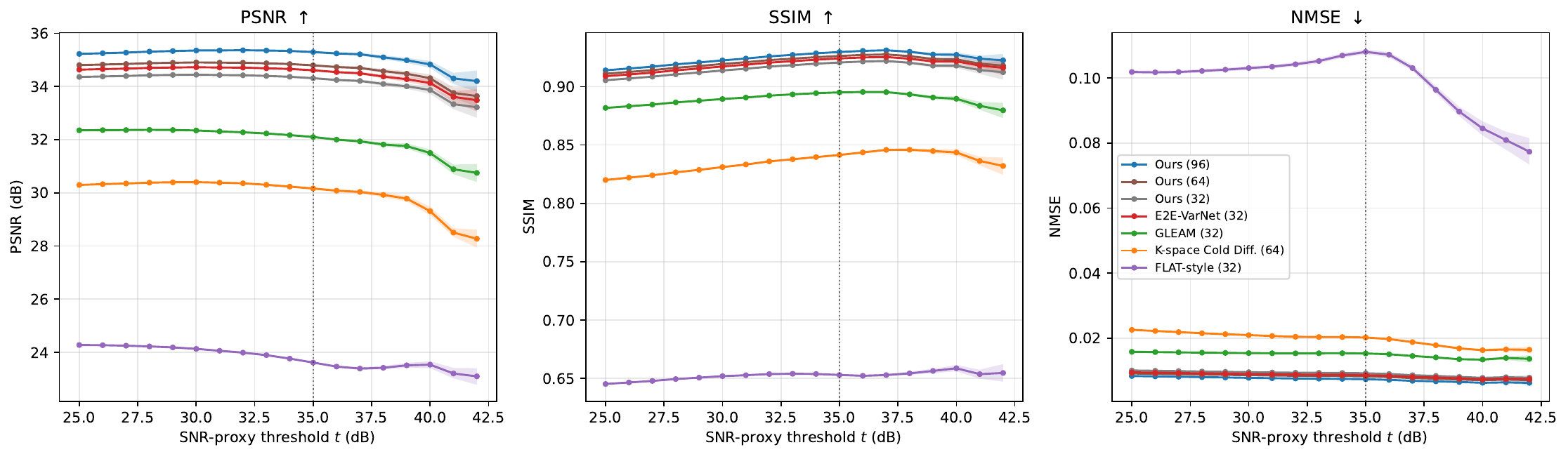}
\caption{\textbf{SNR threshold sensitivity, multi-coil brain $8\times$.} Conventions as in Fig.~\ref{fig:thresh_knee8}.}
\label{fig:thresh_brain8}
\end{figure}

\begin{figure}[t]
\centering
\includegraphics[width=\linewidth]{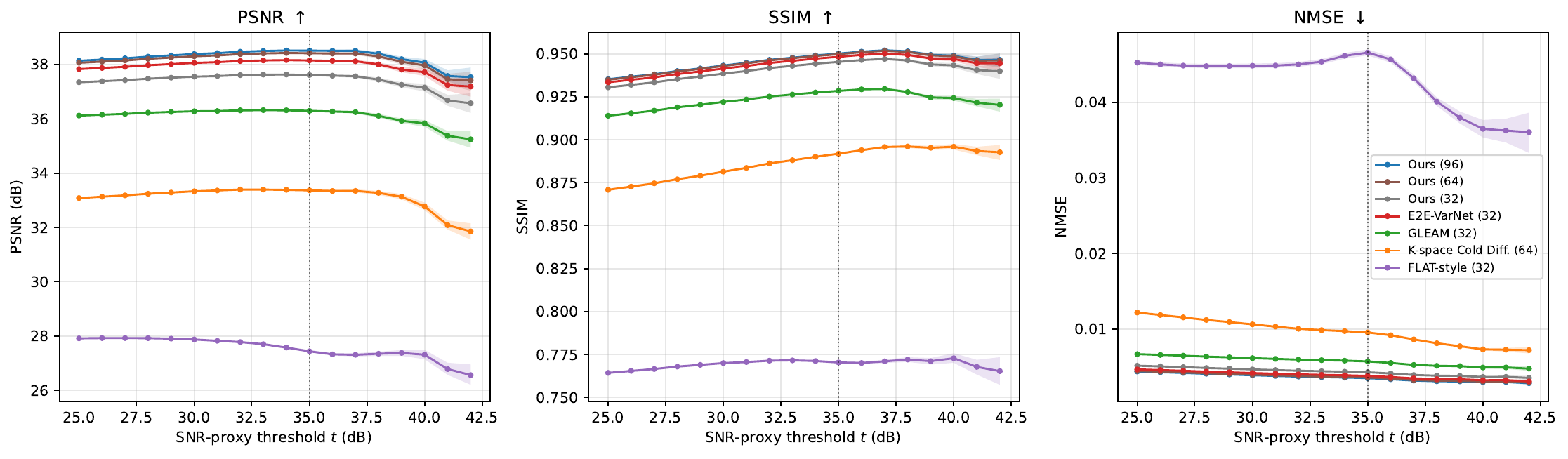}
\caption{\textbf{SNR threshold sensitivity, multi-coil brain $4\times$.} Conventions as in Fig.~\ref{fig:thresh_knee8}.}
\label{fig:thresh_brain4}
\end{figure}

\begin{figure}[t]
    \centering
    \includegraphics[width=0.82\linewidth]{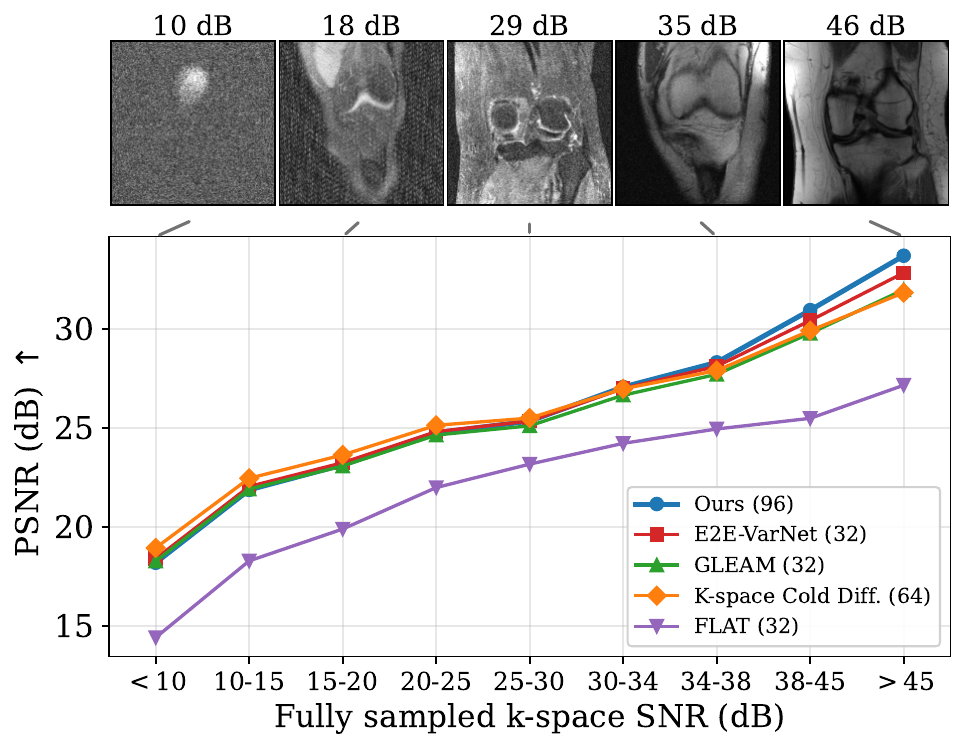}
    \caption{\textbf{Noise sensitivity, single-coil knee $8\times$.}
    Mean PSNR per fully sampled k-space SNR cohort, with reference examples above the curve.
    Fig.~\ref{fig:noise_brain} gives the brain counterpart.}
    \label{fig:noise_knee}
\end{figure}

\begin{figure}[t]
    \centering
    \includegraphics[width=0.82\linewidth]{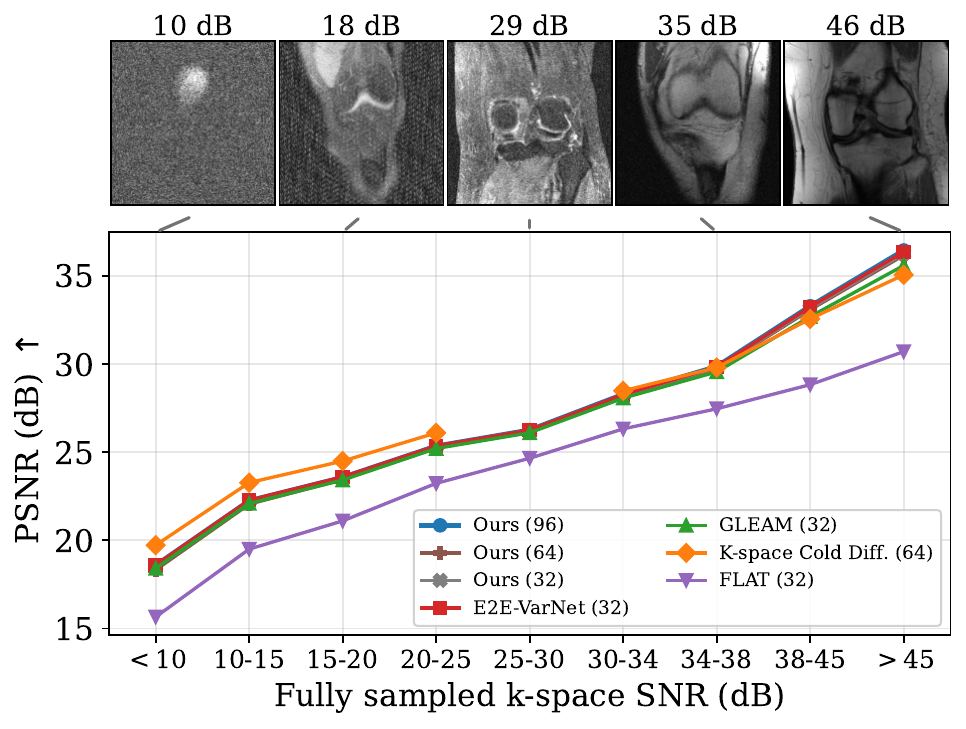}
    \caption{\textbf{Noise sensitivity, single-coil knee $4\times$.}
    As Fig.~\ref{fig:noise_knee}, at $4\times$ acceleration.
    The methods lie closer together than at $8\times$ and the ordering is preserved over the whole range.
    K-space Cold Diffusion again leads only in the noisiest bin.}
    \label{fig:noise_knee4x}
\end{figure}

The SNR-stratified analysis of Tab.~\ref{tab:main} separates the cohorts at $35$\,dB.
Figs.~\ref{fig:thresh_knee8}, \ref{fig:thresh_knee4}, \ref{fig:thresh_brain8} and~\ref{fig:thresh_brain4} sweep this threshold from $25$ to $45$\,dB in steps of $1$\,dB and re-evaluate every method on the cohort SNR${>}t$.
Shaded bands are $95\%$ percentile bootstrap intervals of the cohort mean, obtained by resampling the eligible slices, and the dashed vertical line marks the $35$\,dB cutoff of the main paper.
The relative reconstruction performance of the methods is stable across the threshold range.

Brain is reported without an SNR split (Tab.~\ref{tab:main_brain}), and the distribution of the proxy explains why.
Brain slices occupy a much narrower band than knee slices: the interquartile range is $5.4$\,dB against $8.8$\,dB, $97.6\%$ lie below $40$\,dB where $16.2\%$ of knee slices lie above it, and $17.2\%$ fall below $30$\,dB against $33.0\%$ on knee.
A threshold therefore separates far less of the data on brain, and what it does separate is not acquisition quality: Sec.~\ref{sec:supp_proxy_range} shows the proxy carries essentially no information about difficulty above $30$\,dB there.
Stratifying brain would split a narrow and weakly informative range, which is why the main paper reports it unstratified.

\subsection{Informative range of the SNR proxy}
\label{sec:supp_proxy_range}
The cohort split assumes that our SNR proxy ranks slices by acquisition quality.
We test this assumption by correlating the proxy with the PSNR achieved by Our model, ($C{=}96$, $8\times$) within bands of the proxy itself (Tab.~\ref{tab:proxy_range}).
On knee the proxy is informative over its whole range: the correlation is $0.75$ overall, remains $0.66$ above $30$\,dB, and the mean PSNR rises monotonically from $23.3$ to $31.5$\,dB across the bands.
On brain it is informative only at the noisy end.
Below $25$\,dB the correlation is $0.56$, but between $30$ and $35$\,dB it is $-0.08$ and above $40$\,dB it is $-0.38$.
Over the whole range above $30$\,dB the proxy carries essentially no information about difficulty ($\rho{=}-0.06$). This is a property of the proxy rather than of any particular reconstruction.

\begin{figure}[t]
    \centering
    \includegraphics[width=0.82\linewidth]{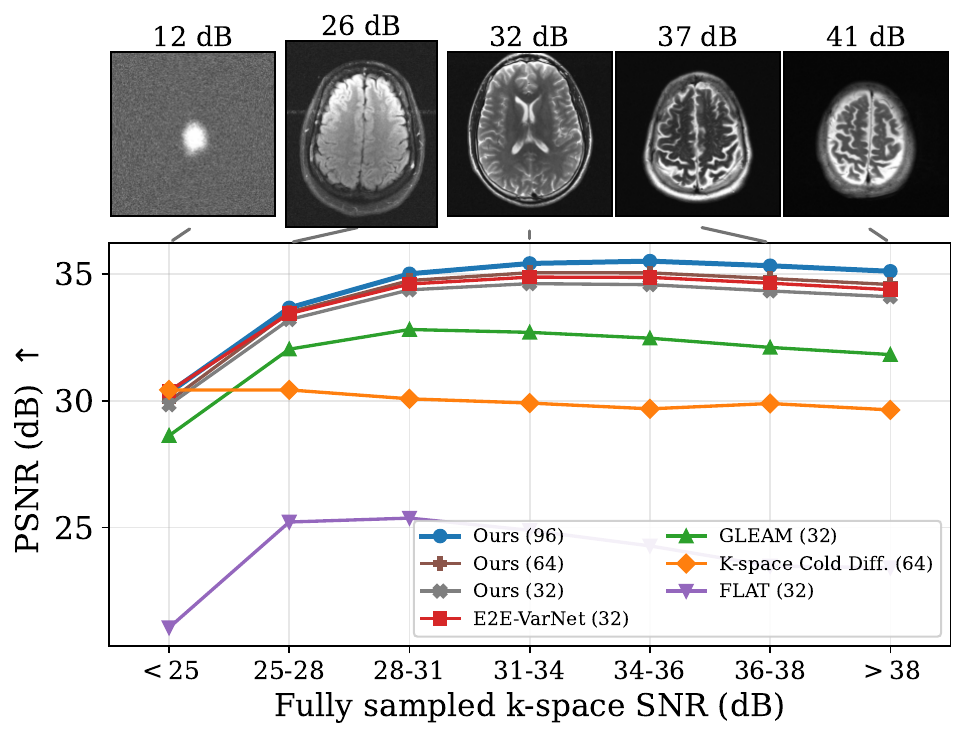}
    \caption{\textbf{Noise sensitivity, multi-coil brain $8\times$.}
    Mean PSNR per fully sampled k-space SNR cohort over the complete validation set ($n{=}21{,}842$ slices, identical for every method, $1075$ to $5158$ per bin), with reference examples above the curve.
    Tab.~\ref{tab:proxy_range} shows that the flatness above $30$\,dB is a property of the proxy on brain rather than of the reconstructions.}
    \label{fig:noise_brain}
\end{figure}

This behavior of the proxy translates to other reconstruction methods. Repeating the analysis for every method of the comparison gives correlations above $30$\,dB between $-0.29$ and $-0.04$ on brain, against $+0.30$ and $+0.66$ on knee, so the proxy fails in the high range on brain, and succeeds on knee, for all of them alike.
This explains the brain curves in Fig.~\ref{fig:noise_brain}, where all methods collapse together below $25$\,dB but the curves are nearly flat above roughly $30$\,dB: on brain the high-SNR cohort separates acquisition type, coil count and coil geometry more than acquisition quality.
Conclusions on brain therefore rest primarily on the results that use no cohort split: the slicewise means over the full validation set in Tab.~\ref{tab:main_brain} and the volumewise evaluation of Sec.~\ref{sec:supp_volumewise}.

\begin{table}[t]
\centering\footnotesize
\caption{\textbf{Where the SNR proxy is informative.}
Spearman correlation between the proxy and the PSNR achieved by a fixed model (Ours, $C{=}96$, $8\times$), within bands of the proxy (knee $n{=}7135$, brain $n{=}21{,}842$).}
\label{tab:proxy_range}
\begin{tabular}{l cc}
\toprule
Proxy band & Brain $\rho$ & Knee $\rho$\\
\midrule
$<25$\,dB      & $+0.56$ & $+0.60$\\
$25$--$30$     & $+0.34$ & $+0.04$\\
$30$--$35$     & $-0.08$ & $+0.10$\\
$35$--$40$     & $+0.12$ & $+0.37$\\
$>40$          & $-0.38$ & $+0.35$\\
\midrule
overall        & $+0.17$ & $+0.75$\\
above $30$\,dB & $-0.06$ & $+0.66$\\
\bottomrule
\end{tabular}%

\end{table}

\begin{figure}[t]
    \centering
    \includegraphics[width=0.82\linewidth]{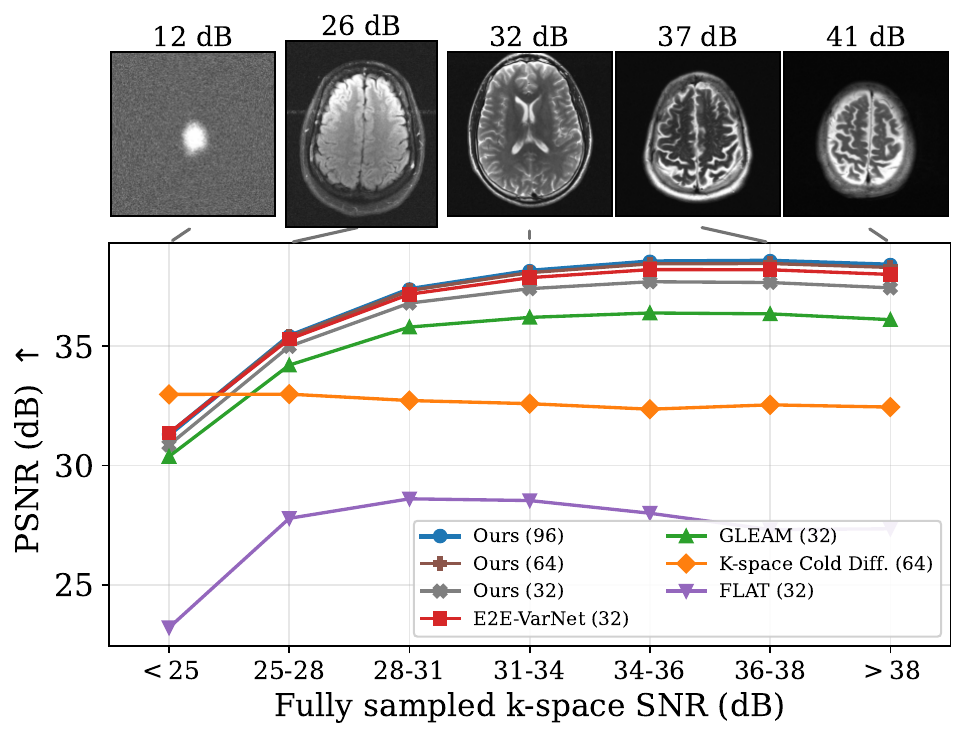}
    \caption{\textbf{Noise sensitivity, multi-coil brain $4\times$.}
    As Fig.~\ref{fig:noise_brain}, at $4\times$ acceleration ($n{=}21{,}842$ slices, identical slices for every method).
    The curves are flatter still than at $8\times$: with the inverse problem better constrained, acquisition quality separates the methods less.}
    \label{fig:noise_brain4x}
\end{figure}

\subsection{SNR proxy relation to slice position}
\label{sec:supp_slicepos}
Volumewise metrics would be sufficient to account for low-SNR slices if they were confined to the near-empty regions at the ends of each volume, but we find that low SNR also occurs in central slices.
We therefore relate the proxy to the position of each slice within its volume, over all $7{,}135$ single-coil knee validation slices (Tab.~\ref{tab:slicepos}).
Slices in the central half of a volume, where anatomy fills the field of view, are cleaner on average than those in the outer quarters, but noise is not confined to the volume ends: $25.8\%$ of the central slices fall below $30$\,dB and $4.4\%$ below $25$\,dB, and the noisiest central slice has $13.5$\,dB.
Of all slices below $30$\,dB, $38\%$ are central, and the correlation between the proxy and the distance from the volume center is only $-0.25$.
Low SNR is instead largely a property of the acquisition as a whole: volume medians span $22.7$ to $46.3$\,dB, $49$ of the $199$ volumes have a median below $30$\,dB, and $31$ fall below $30$\,dB on every slice.
Half of the variance of the proxy ($51\%$) lies between volumes rather than within them.
A volumewise mean over such a volume is therefore itself confounded by low SNR.

\begin{table}[t]
\centering\footnotesize
\caption{\textbf{SNR proxy by slice position, single-coil knee.}
All $7{,}135$ validation slices, split by position within the volume: central is the middle half, edge the outer quarter at each end.}
\label{tab:slicepos}
\begin{tabular}{l cc}
\toprule
 & Central & Edge\\
\midrule
Slices & $3{,}503$ & $3{,}632$\\
Median proxy (dB) & $33.5$ & $32.6$\\
Below $30$\,dB & $25.8\%$ & $39.9\%$\\
Below $25$\,dB & $4.4\%$ & $20.9\%$\\
Minimum (dB) & $13.5$ & $-6.7$\\
\bottomrule
\end{tabular}%

\end{table}

\section{Method analyses}
\subsection{Components of the compact backward}
\label{sec:supp_bwablation}
Sec.~\ref{sec:exp_efficiency} of the main paper considers the activation-light backward as a single mechanism.
Tab.~\ref{tab:bwablation} decomposes it into its three components, enabling them one at a time with everything else held fixed.
All four levels run with the suffix parameters frozen, so the set of updated parameters is identical throughout and only the storage scheme varies.

The levels differ in what each layer has to keep from the forward pass in order to backpropagate through it.
At L1 the convolutions save only their weights, which are stored as parameters in any case. 
The difference to the original implementation is that the input activations are never stored as the weight gradients are not needed for activation-light backpropagation.
At L2 the LeakyReLU layers exploit the fact that their backward needs only the sign of the input, and keep a one-bit mask per element, packed eight to a byte, in place of the activation.
At L3 the instance normalizations retain the standard forward but treat their statistics as constants when differentiating, which requires only a scalar per channel and sample (and optionally the normalization multiplying factor), instead of the activation.
L1 and L2 reproduce the reverse mode gradient exactly, whereas the frozen statistics of L3 discard the terms that flow through the dependence of the mean and variance on the input, the only approximation in the method.

No single component dominates the memory saving.
The increments are $1.7$, $1.8$ and $2.0$\,GB.
The two exact levels alone reduce the peak from $6.6$ to $3.1$\,GB, and the frozen-statistics normalization accounts for the remaining step to $1.1$\,GB, the operating point of Tab.~\ref{tab:isolate}.

The saving also reduces the training time.
Step time falls from $154$ to $134$\,ms at L1, where discarding the stored inputs also removes the weight gradient computation, and the remaining levels leave it essentially unchanged (at $134$\,ms for the complete backward).

\begin{table}[t]
\centering\footnotesize
\caption{\textbf{Componentwise ablation of the activation-light backward.}
Single-coil knee $8\times$, $T{=}8$, $C{=}32$, single A6000, batch $1$.
Levels are cumulative.
Peak memory is the maximum over the sampled cascade index and step time the mean over it.}
\label{tab:bwablation}
\begin{tabular}{l cc c}
\toprule
Suffix backward component & \makecell{Peak\\(GB)$\downarrow$} & \makecell{Time\\(ms)} & Gradients\\
\midrule
L0\quad standard autograd & 6.61 & 154 & exact\\
L1\quad + conv/linear adjoint & 4.93 & 134 & exact\\
L2\quad + LeakyReLU sign masks & 3.10 & 137 & exact\\
L3\quad + frozen-statistics norm & 1.14 & 134 & approx.\\
\bottomrule
\end{tabular}%

\end{table}

\subsection{Broader gradient fidelity distribution}
\label{sec:supp_jacobian_broad}
\begin{figure*}[t]
\centering
\includegraphics[width=0.85\textwidth]{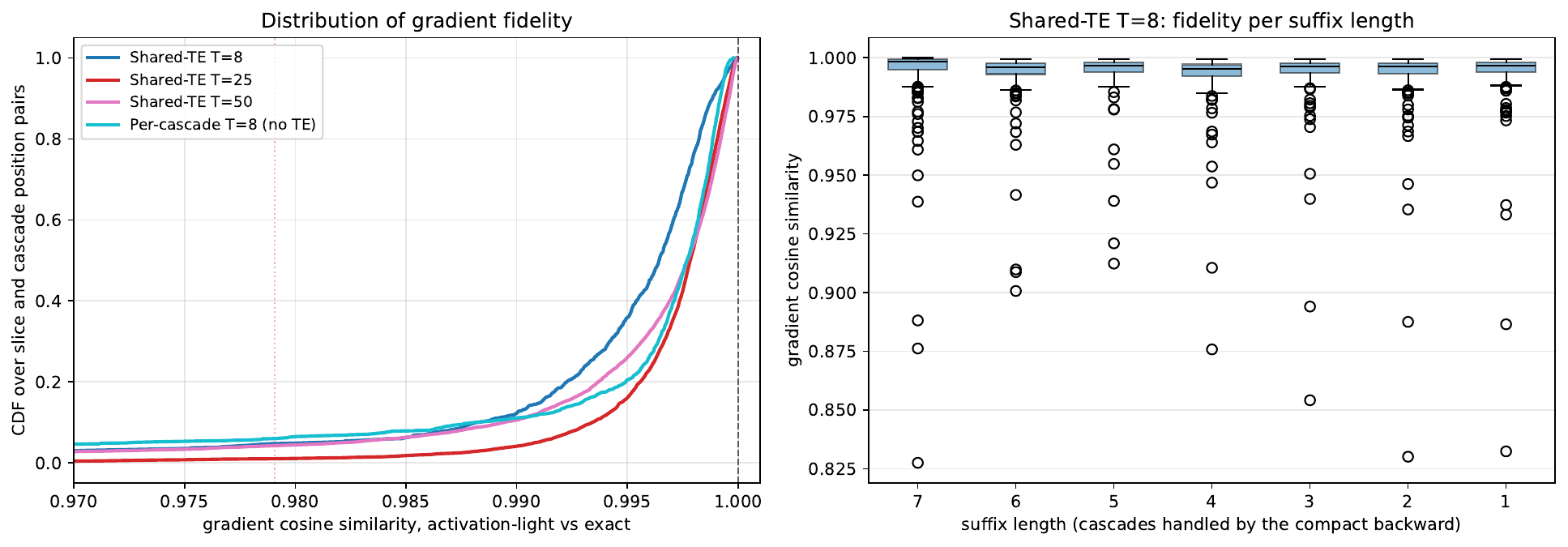}
\caption{\textbf{Gradient fidelity distribution, single-coil knee $8\times$, $C{=}32$.}
The quantity compared is the gradient the suffix returns to the output of the sampled cascade, computed with the compact backward and with exact autograd.
Left: CDF of their cosine similarity over $n{=}200$ slices and every sampled cascade position, with dotted vertical lines at the $1$st percentile of each model.
Eight of the $5{,}000$ measurements at $T{=}50$ exceed memory in the exact reference and are excluded.
Right: box plot per suffix length for the shared time-embedded model at $T{=}8$.}
\label{fig:jac_broad}
\end{figure*}

The gradient fidelity check in Sec.~\ref{sec:exp_fidelity} of the main paper reports the median cosine similarity over $n{=}200$ high-SNR knee slices at $T{=}8$.
Here we give the full distribution behind it, over the $200$ slices and all $7$ sampled cascade positions per slice ($1{,}400$ paired measurements per model), for both the shared time-embedded network of the paper and the per-cascade variant of Tab.~\ref{tab:abl_vol}.
Fig.~\ref{fig:jac_broad} shows the CDF over all (slice, cascade position) pairs and the distribution per suffix length.

The median cosine similarity is $\approx\!0.996$ for both architectures, but the tail is wider: $99\%$ of the pairs lie above $0.91$ for the shared network and above $0.82$ for the per-cascade variant, with minima of $0.83$ and $0.27$ respectively.
The shared time-embedded architecture thus has systematically tighter suffix Jacobian fidelity than the per-cascade variant.

Fidelity does not deteriorate as the chain lengthens.
Repeating the measurement at $T{=}25$ and $T{=}50$ over the same $200$ slices and every sampled cascade position ($4{,}800$ and $4{,}992$ pairs) gives a median of $0.998$ at both depths, against $0.996$ at $T{=}8$, with $99\%$ of the pairs above $0.98$ and $0.93$ respectively.
The suffix that the compact backward traverses is therefore longer without being less faithful in the bulk of the distribution.
The extreme tail is depth dependent, however: the smallest cosine observed is $0.91$ at $T{=}25$ but $0.42$ at $T{=}50$, against $0.83$ at $T{=}8$.

\subsection{Iteration time of activation checkpointing}
\label{sec:supp_ckpt}
Activation checkpointing (Chen et al.\ 2016, cited in the main paper) discards the activations of each cascade during the forward pass and recomputes them during backpropagation.
Its gradients are exact, so reconstruction quality equals that of standard E2E-VarNet by construction and the entire cost falls on the step time.
Sec.~\ref{sec:supp_ckptgran} examines what the scheme buys; here we quantify what it costs as the chain lengthens.

Recomputing the discarded activations costs a constant factor rather than a growing one.
Against full backpropagation it is $1.3$ to $1.4\times$ at every depth we measure: $215\rightarrow293$\,ms at $T{=}8$, $679\rightarrow901$\,ms at $T{=}25$ and $1367\rightarrow1805$\,ms at $T{=}50$.
That factor, however, multiplies a cost that itself grows with the chain, so the absolute penalty compounds.
Relative to our method, checkpointing is $159$\,ms slower per step at $T{=}8$, $537$\,ms slower at $T{=}25$ and $1.09$\,s slower at $T{=}50$, a ratio of $2.2$ to $2.5\times$ throughout.
Scaling width behaves the same way: at the flagship configuration $C{=}96$, $T{=}15$ checkpointed E2E-VarNet needs $2.08$\,s per step against $0.94$\,s for our method.
Checkpointing is therefore a sound exact gradient option wherever the time budget allows it, but it is slower than both full backpropagation and our method at every depth in Tab.~\ref{tab:ckptseg}.

\subsection{Checkpointing granularity}
\label{sec:supp_ckptgran}

Every checkpointing number in this work checkpoints at the granularity of one cascade: the input of each cascade is stored and its interior activations are recomputed during the backward pass.
A finer alternative nests a second level of checkpoints inside the U-Net blocks of each cascade, so that the block interiors are recomputed as well.
Tab.~\ref{tab:ckptseg} measures both against full backpropagation and against our method.

Nesting does lower the peak further, by ${\approx}0.3$\,GB at every depth ($1.26\rightarrow0.96$\,GB at $T{=}8$ and $3.77\rightarrow3.47$\,GB at $T{=}50$), which is the difference between holding the activations of one cascade and those of one block.
It pays for this in recomputation, $25$ to $31\%$ more time per iteration ($293\rightarrow385$\,ms at $T{=}8$ and $1805\rightarrow2249$\,ms at $T{=}50$), since the interiors are now recomputed twice rather than once.
This is the recursive variant of Chen et al., whose computation grows with the nesting depth.

Neither granularity changes how checkpointed memory scales with depth, and the reason is visible in the last two columns of Tab.~\ref{tab:ckptseg}.
Peak memory is the sum of the activations and of the weights and gradients, and checkpointing acts only on the first of these.
On that term it works: per cascade checkpointing holds the activations at $0.80$ to $0.88$\,GB from $T{=}8$ to $T{=}50$, and nesting takes a further $0.3$\,GB off.
The second term is beyond the reach of any activation scheme and grows linearly, because E2E-VarNet keeps a distinct network per cascade: $0.46$, $1.44$ and $2.89$\,GB at $T{=}8$, $25$ and $50$.
By $T{=}50$ it is $83\%$ of the nested checkpointing peak, so a finer granularity has almost nothing left to remove.

Sharing one network across the cascades is what removes the growing term, which stays at $0.06$\,GB at every depth.
The two strategies therefore separate as the chain lengthens: from $T{=}8$ to $T{=}50$ our peak rises by $0.18$\,GB and that of nested checkpointing by $2.51$\,GB, so nesting is $0.14$\,GB ahead of us at $T{=}8$ and $2.19$\,GB behind at $T{=}50$.
Ours is also the only scheme in Tab.~\ref{tab:ckptseg} faster than full backpropagation.

\begin{table}[t]
\centering\footnotesize
\caption{\textbf{Checkpointing schemes}, single-coil knee, $C{=}32$, one A6000 at batch $1$.
Per cascade checkpointing stores the input of each cascade and is the setting used everywhere else in the paper; nested adds a second level inside the U-Net blocks.
Peak memory is split into the weights and gradients, which stay resident throughout, and the activations, which are the remainder.}
\label{tab:ckptseg}
\setlength{\tabcolsep}{3.5pt}%
\begin{tabular}{l cccc}
\toprule
Scheme & \makecell{Peak\\(GB)$\downarrow$} & \makecell{weights\\$+$grads} & \makecell{activations} & \makecell{Time\\(ms)}\\
\midrule
\multicolumn{5}{l}{\emph{$T{=}8$}}\\
\quad full backprop.     & $5.94$ & $0.46$ & $5.48$ & $215$\\
\quad ckpt.\ per cascade & $1.26$ & $0.46$ & $0.80$ & $293$\\
\quad ckpt.\ nested      & $0.96$ & $0.46$ & $0.50$ & $385$\\
\quad \emph{Ours}        & $1.10$ & $0.06$ & $1.04$ & $134$\\
\midrule
\multicolumn{5}{l}{\emph{$T{=}25$}}\\
\quad full backprop.     & $18.33$ & $1.44$ & $16.89$ & $679$\\
\quad ckpt.\ per cascade & $2.27$  & $1.44$ & $0.83$  & $901$\\
\quad ckpt.\ nested      & $1.98$  & $1.44$ & $0.54$  & $1237$\\
\quad \emph{Ours}        & $1.18$  & $0.06$ & $1.12$  & $364$\\
\midrule
\multicolumn{5}{l}{\emph{$T{=}50$}}\\
\quad full backprop.     & $36.54$ & $2.89$ & $33.65$ & $1367$\\
\quad ckpt.\ per cascade & $3.77$  & $2.89$ & $0.88$  & $1805$\\
\quad ckpt.\ nested      & $3.47$  & $2.89$ & $0.58$  & $2249$\\
\quad \emph{Ours}        & $1.28$  & $0.06$ & $1.22$  & $719$\\
\bottomrule
\end{tabular}%

\end{table}

\subsection{Peak memory scaling curves}
\label{sec:supp_memcurves}
\begin{figure*}[t]
\centering
\begin{subfigure}{0.46\textwidth}
\centering
\includegraphics[width=\linewidth]{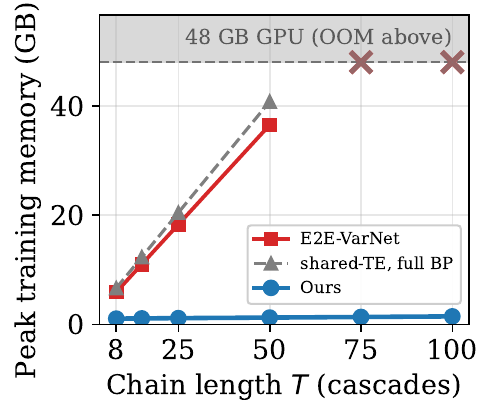}
\caption{Peak memory against chain length ($C{=}32$).}
\label{fig:mem_depth}
\end{subfigure}\hspace{0.04\textwidth}
\begin{subfigure}{0.46\textwidth}
\centering
\includegraphics[width=\linewidth]{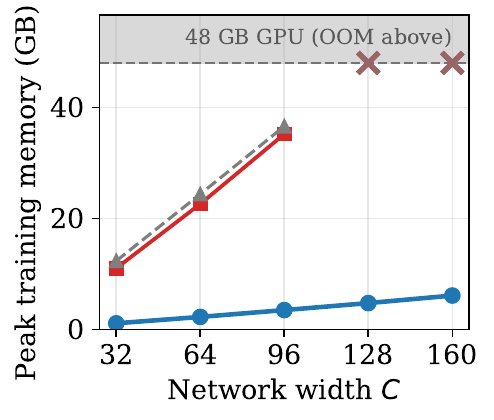}
\caption{Peak memory against network width ($T{=}8$).}
\label{fig:mem_width}
\end{subfigure}
\caption{\textbf{Peak training memory}, batch $1$ on one A6000.
Crosses mark configurations that exceed the $48$\,GB budget.}
\label{fig:mem_curves}
\end{figure*}

Sec.~\ref{sec:exp_efficiency} of the main paper reports peak training memory at selected depths and widths, and Fig.~\ref{fig:mem_curves} plots the full curves.
Along both axes E2E-VarNet grows steeply and exceeds the $48$\,GB budget of a single A6000, while our peak stays nearly flat.
A shared time-embedded backbone trained with full backpropagation tracks E2E-VarNet and runs out of memory at the same point, which locates the saving in the activation handling rather than in the architecture.

\section{Implementation details}
\subsection{U-Net implementation and cascade conditioning}
\label{sec:supp_teunet}
All unrolled methods in this paper (E2E-VarNet, GLEAM, FLAT-style, and ours) use the same U-Net backbone as the cascade regularizer: the standard fastMRI U-Net with $4$ resolution levels, stage widths $C\rightarrow 2C\rightarrow 4C\rightarrow 8C$ with a $16C$ bottleneck, two $3{\times}3$ convolutions per block, each followed by instance normalization and LeakyReLU (with a negative slope of $0.2$), $2{\times}2$ average pool downsampling, transposed convolution upsampling, and a final $1{\times}1$ projection to two channels.
The baselines instantiate one such network per cascade.
Our model instead shares a single network across all cascades and conditions it on the cascade index.
We specify the conditioning here because it carries the entire gain over independent per-cascade networks.

\noindent\textbf{Conditioning path.}
The cascade index $t\in\{0,\dots,T-1\}$ is mapped to a sinusoidal position embedding of width $C$, as is standard in diffusion models (Ho et al.\ 2020, cited in the main paper), and passed through a two-layer MLP ($C\rightarrow 6C\rightarrow 6C$, GELU).
This yields one embedding vector of width $6C$ per sample ($192$ at $C{=}32$, $576$ at $C{=}96$), which is shared by every block of the network.

\noindent\textbf{Injection points.}
Each double-convolution block receives, between its two convolutions, a per-channel additive shift obtained from the shared embedding through a SiLU and a linear map $6C\rightarrow C_{\text{out}}$.
The conditioning is therefore an additive channel-wise bias rather than a modulation of the weights: the operator applied at each cascade is the same function of the image, displaced by a cascade-dependent offset in feature space.
There are $9$ injection points (four in the encoder, one at the bottleneck and four in the decoder).
The upsampling path and the output projection are not conditioned.

\noindent\textbf{Cost of conditioning.}
The conditioning path (the shared embedding MLP plus the nine per-block linear maps) accounts for $327$k of the $8.084$M parameters at $C{=}32$ and for $2.94$M of $72.740$M at $C{=}96$, close to $4\%$ in both cases.
Since the network is shared by all $T$ cascades, it is the entire model: at $T{=}8$, $C{=}32$ the reconstruction chain has $8.08$M parameters in total, against $62.05$M for the same chain built from independent per-cascade U-Nets.

\noindent\textbf{Initialization.}
The final projection is scaled by $1/\sqrt{T}$ at initialization, so that the superposition of $T$ cascade updates keeps the output norm approximately independent of the chain length.

\subsection{Measurement protocol}
\label{sec:supp_measurement}
All training and profiling use full fp32 precision without mixed precision.
Peak memory is \texttt{torch.cuda.max\_memory\_allocated} measured over a complete training step (forward and backward) and includes parameters, gradients and activations.
Optimizer state is excluded for all methods alike, since it depends on the parameter count rather than on the training scheme under study.
Batch size $1$ isolates the activation footprint of a single sample.
Activation memory scales approximately linearly with the batch size for every method compared here, so the relative comparisons do not depend on it.
Training itself uses an effective batch of $8$, accumulated from such microbatches.
All measurements use a single NVIDIA RTX~A6000 ($48$\,GB) at $640{\times}372$ resolution.

\section{Additional qualitative examples}
\label{sec:supp_qualitative}
Figs.~\ref{fig:gallery_knee} and~\ref{fig:gallery_brain} extend the qualitative comparison of the main paper (Fig.~1) with further validation slices for both anatomies.
Figs.~\ref{fig:gallery_knee4x} and~\ref{fig:gallery_brain4x} repeat the comparison at $4\times$ on the same slices.
All methods lie closer together at $4\times$, where the inverse problem is better conditioned, and the differences that remain are concentrated in finer anatomical structures.

\begin{figure*}[p]
\centering
\begin{subfigure}{\textwidth}
\centering
\includegraphics[width=\textwidth]{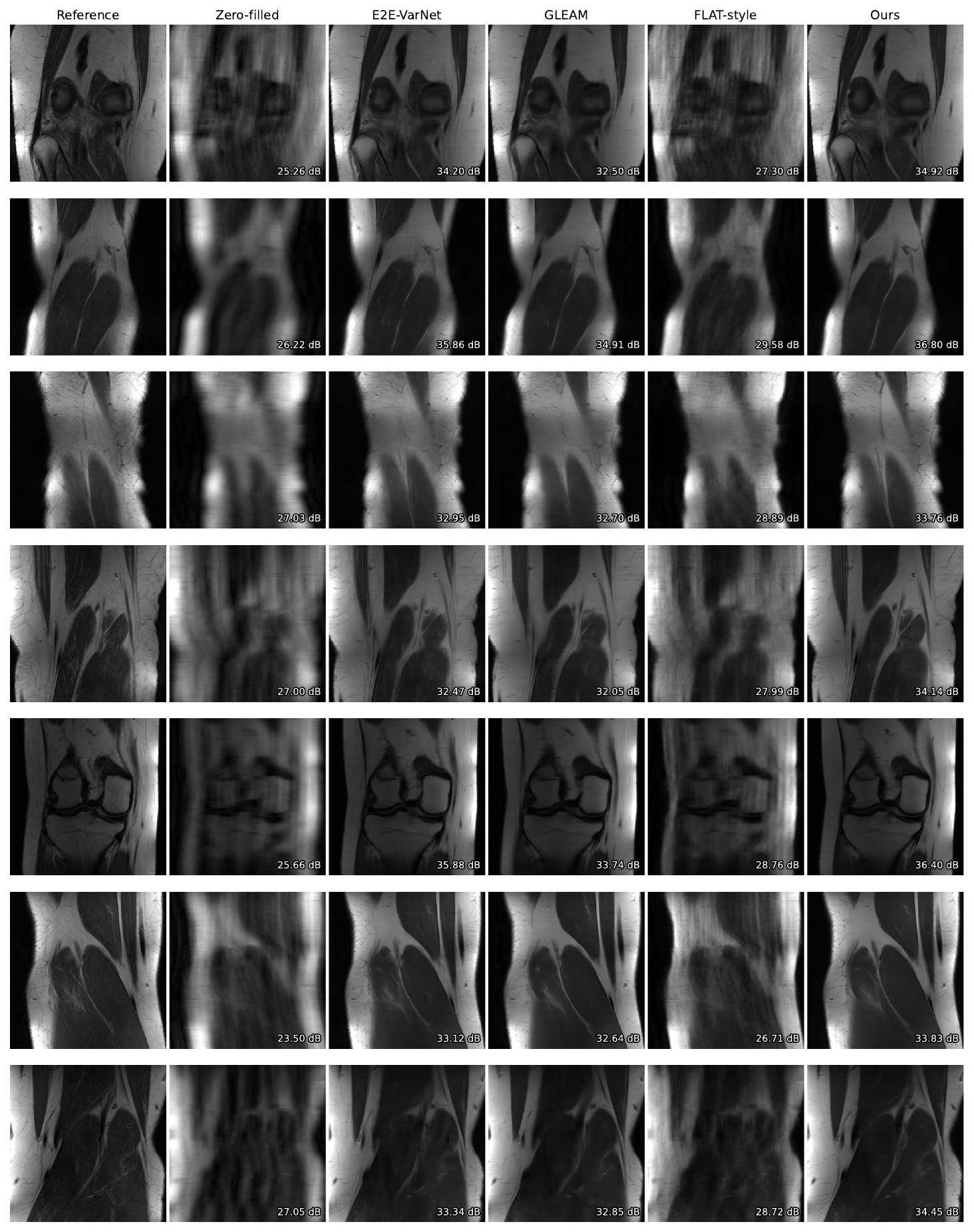}
\caption{$8\times$.}
\label{fig:gallery_knee}
\end{subfigure}
\caption{\textbf{Additional single-coil knee reconstructions.}
Rows are validation slices, columns the compared methods and the fully sampled reference.
The two accelerations show the same slices.}
\end{figure*}

\begin{figure*}[p]\ContinuedFloat
\centering
\begin{subfigure}{\textwidth}
\centering
\includegraphics[width=\textwidth]{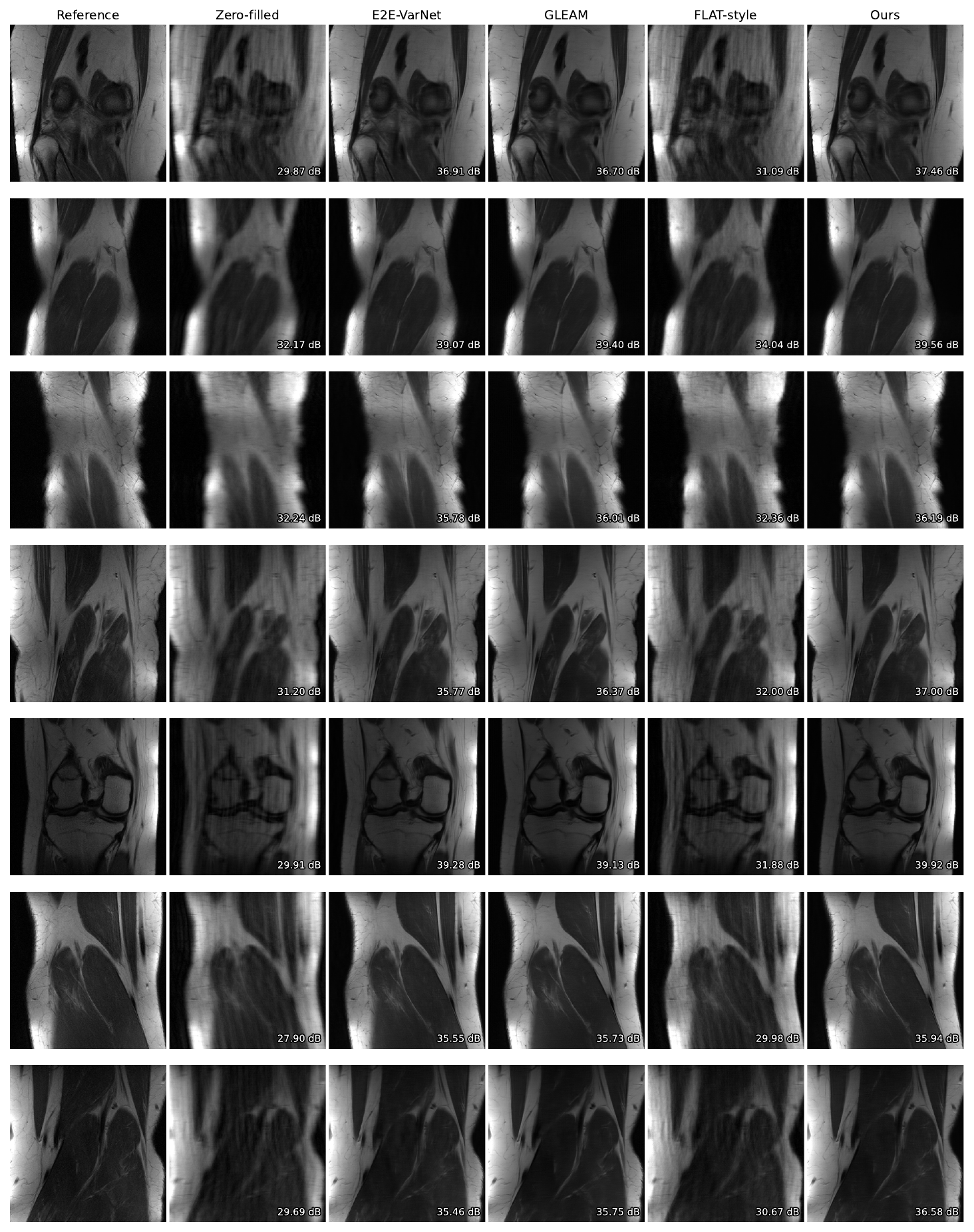}
\caption{$4\times$.}
\label{fig:gallery_knee4x}
\end{subfigure}
\caption{Additional single-coil knee reconstructions (continued).}
\end{figure*}

\begin{figure*}[p]
\centering
\begin{subfigure}{\textwidth}
\centering
\includegraphics[width=\textwidth]{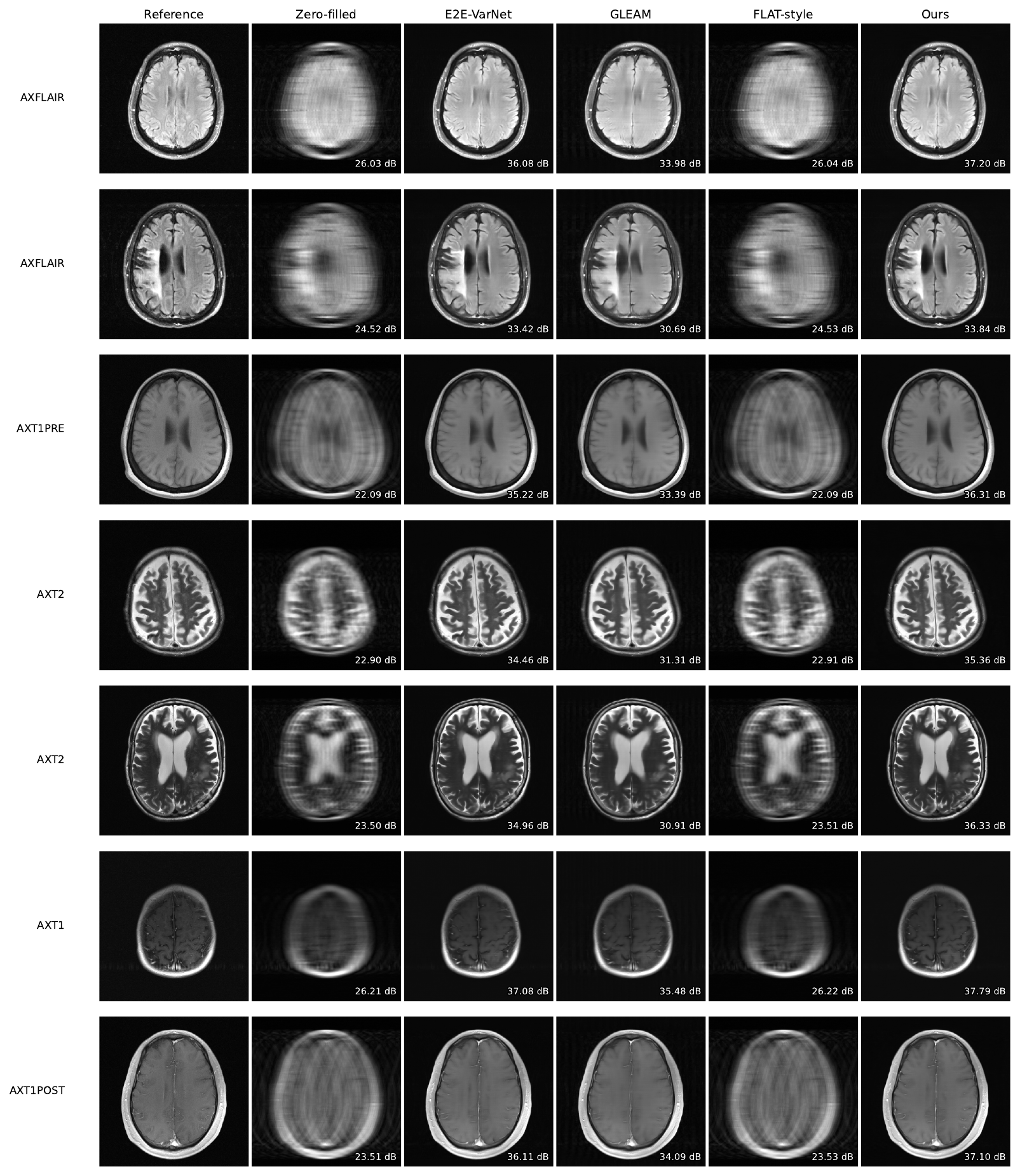}
\caption{$8\times$.}
\label{fig:gallery_brain}
\end{subfigure}
\caption{\textbf{Additional multi-coil brain reconstructions.}
Rows are validation slices across contrasts, columns as in Fig.~\ref{fig:gallery_knee}.
The two accelerations show the same slices.}
\end{figure*}

\begin{figure*}[p]\ContinuedFloat
\centering
\begin{subfigure}{\textwidth}
\centering
\includegraphics[width=\textwidth]{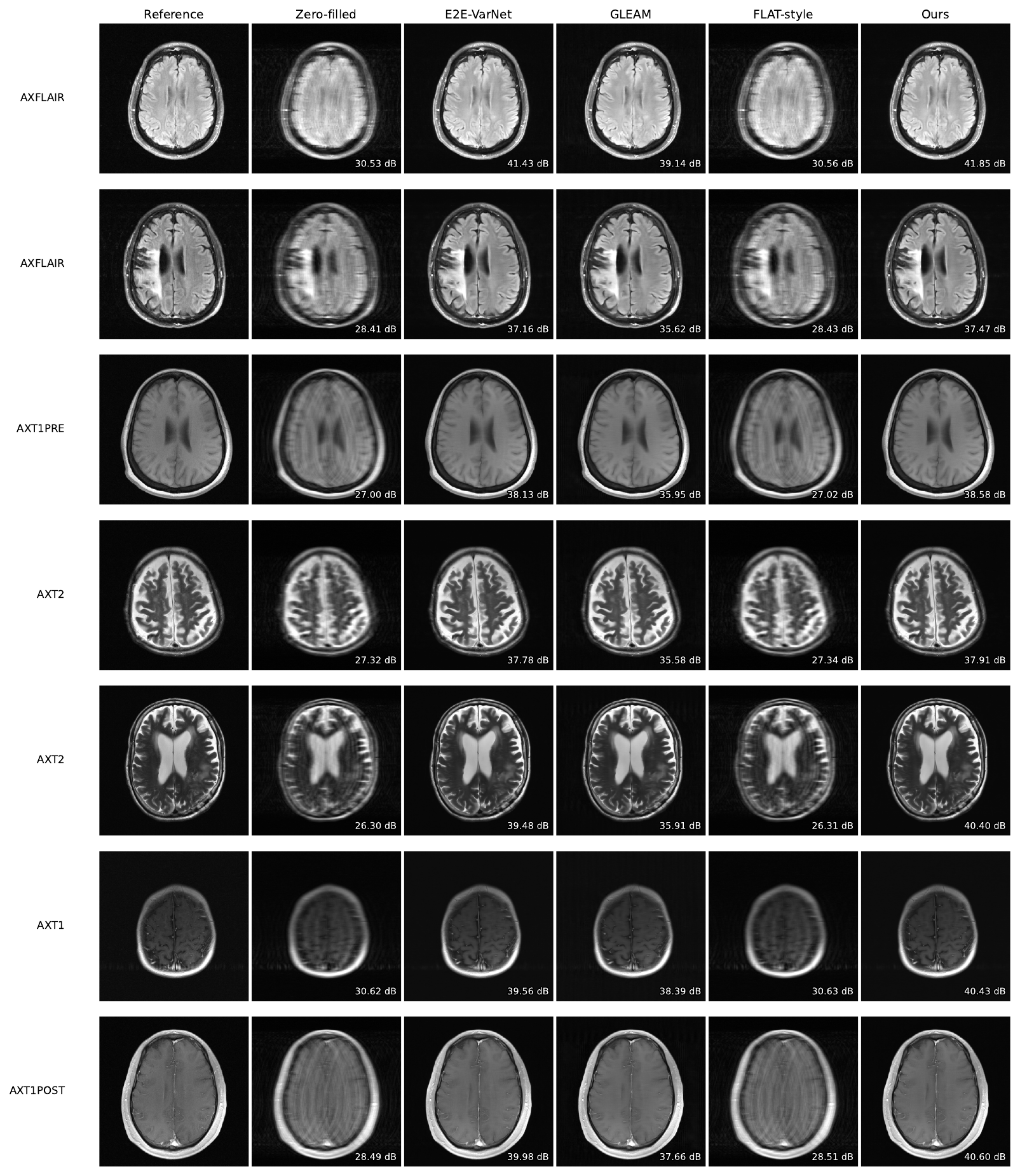}
\caption{$4\times$.}
\label{fig:gallery_brain4x}
\end{subfigure}
\caption{Additional multi-coil brain reconstructions (continued).}
\end{figure*}

\end{document}